\documentclass[preprint,12pt,authoryear]{elsarticle}

\usepackage[normalem]{ulem}
\usepackage{amsmath,amssymb, amsthm}
\usepackage{multirow}
\usepackage{booktabs}
\usepackage{mathtools}
\usepackage{hyperref}
\usepackage{xcolor}
\usepackage{soul}
\usepackage[prependcaption,textsize=footnotesize]{todonotes}
\usepackage{xargs}
\usepackage{fancyhdr}
\usepackage{epstopdf}
\usepackage{cite}
\usepackage{url}

\journal{Neural Networks}

\sethlcolor{lightgray}
\newcommandx{\trashit}[2][1=]{{\color{red}#2}\todo[linecolor=red,backgroundcolor=red!25,bordercolor=red, inline]{#1}}
\newcommandx{\rewrite}[2][1=]{{\color{green}#2}\todo[linecolor=blue,backgroundcolor=blue!25,bordercolor=blue, inline]{#1}}

\newcommand{\opush}{\odot}
\newcommand{\opop}{\ominus}
\newcommand{\ra}[1]{\renewcommand{\arraystretch}{#1}}
\newcommand{\pss}[3]{\prescript{\text{\tiny#1}}{}{#2}_\text{#3}}

\renewcommand{\vec}[1]{\mbox{\boldmath$#1$}}

\begin{document}

\begin{frontmatter}





\title{A modular architecture for transparent computation\\ in Recurrent Neural Networks}

\author[pu]{Giovanni S. Carmantini\corref{cor1}}
\ead{giovanni.carmantini@gmail.com}

\author[bc]{Peter beim Graben}

\author[is]{Mathieu Desroches}

\author[pu]{Serafim Rodrigues}
\cortext[cor1]{Corresponding author}

\address[pu]{School of Computing and Mathematics, Plymouth
    University,\\ Plymouth, United Kingdom}
\address[bc]{Bernstein Center for Computational Neuroscience Berlin,\\
  Humboldt-Universit\"{a}t zu Berlin, Berlin, Germany}
\address[is]{Inria Sophia-Antipolis M\'{e}diterran\'{e}e, Valbonne, France}

\begin{abstract}
  Computation is classically studied in terms of automata, formal languages and algorithms; yet, the relation between neural dynamics and symbolic representations and operations is still unclear in traditional eliminative connectionism. Therefore, we suggest a unique perspective on this central issue, to which we would like to refer as to transparent connectionism, by proposing accounts of how symbolic computation can be implemented in neural substrates. In this study we first introduce a new model of dynamics on a symbolic space, the versatile shift, showing that it supports the real-time simulation of a range of automata. We then show that the G\"odelization of versatile shifts defines nonlinear dynamical automata, dynamical systems evolving on a vectorial space. Finally, we present a mapping between nonlinear dynamical automata and recurrent artificial neural networks. The mapping defines an architecture characterized by its granular modularity, where data, symbolic operations and their control are not only distinguishable in activation space, but also spatially localizable in the network itself, while maintaining a distributed encoding of symbolic representations. The resulting networks simulate automata in real-time and are programmed directly, in absence of network training. To discuss the unique characteristics of the architecture and their consequences, we present two examples: i) the design of a Central Pattern Generator from a finite-state locomotive controller, and ii) the creation of a network simulating a system of interactive automata that supports the parsing of garden-path sentences as investigated in psycholinguistics experiments.
\end{abstract}


\end{frontmatter}


\section{Introduction}
The relation between symbolic computation and neural dynamics is one of the most pertinent problems in computational neuroscience, artificial intelligence, and cognitive science. On the one hand, symbolic computation is generically codified in terms of production systems, formal languages, algorithms and automata \citep{HopcroftUllman79}. On the other hand, neural dynamics in artificial neural networks (ANN) is described by nonlinear evolution laws \citep{HertzKroghPalmer91}. Approaches to connect these different realms of research go back to the seminal paper of \citet{mcculloch1943logical} on networks of idealized two-state neurons that behave as logic gates. Furthermore, fundamental work by \citet{kleene1956neural} and \citet{minsky1967computation} demonstrated the equivalence between such networks and finite-state automata, and thus digital computers (which are essentially large-scale networks of logic gates). Later examples for connectionist modeling of symbolic computation are the speech perception and production models \emph{TRACE} by \citet{McClellandElman86} and \emph{NETtalk} by \citet{SejnowskiRosenberg87}. A further important step was achieved by Elman when introducing simple recurrent networks (SRN) as prediction devices for letters in words \citep{Elman90} and syntactic categories in sentences \citep{Elman95}. SRN found a number of successful applications in linguistics and cognitive science \citep{TaborJulianoTanenhaus97, ChristiansenChater99b, LawrenceGilesFong00, FarkasCrocker08} where formal grammars have been employed for the generation of training sets. After training, grammatical relations emerged in the connectivity and activation patterns of the network's hidden layer which could be examined through clustering and principal component analysis (PCA).

A key problem of this and similar approaches based on \emph{eliminative connectionism} \citep{Blutner11}, a theoretical stance aiming at the elimination of symbolic representations in connectionist models, is that the emerging representations, while comparable in a metric space through empirical methods such as clustering or PCA, do not allow inferences about the syntactic or structural relationships of the symbolic training data. This is even more the case with contemporary deep-learning \citep{Deng14, BengioCourvilleVincent13}, and reservoir computing approaches featuring large networks of randomly and recurrently connected nonlinear units \citep{dominey1995complex, jaeger2001echo, MaassNatschlagerMarkram02, steil2004backpropagation}. For that reason, another branch of research, which we may call \emph{transparent connectionism}, has been developed in the framework of vector symbolic architectures (VSA) \citep{Mizraji89, Smolensky90, SmolenskyLegendre06a, SmolenskyLegendre06b, Gayler06, GaylerLevyBod10, graben2009inverse}. Here, one explicitly starts with the symbolic data structures and processes, which are first decomposed into so-called filler-role bindings and then used to create vectorial images through tensor product representations \citep{Smolensky90, graben2009inverse}. These serve as training patterns for subsequent connectionist modeling. In contrast to eliminative connectionism where representations that emerge during training are to a great extent opaque, representations in VSAs are completely transparent as they can be resolved in each step of the encoding procedure.
Depending on the structure of the chosen vector space one arrives at different kinds of integrated connectionist/symbolic architectures (ICS) \citep{Smolensky90, SmolenskyLegendre06a, SmolenskyLegendre06b}: G\"odelizations for one-dimensional representations in the field of real numbers, proper vectorial representations for finite-dimensional vector spaces, and functional representations for infinite-dimensional vector spaces \citep{graben2009inverse}.
Importantly, \citet{siegelmann_math_1991, siegelmann_computational_1995} used a combination of G\"odelization and localist finite-dimensional representation to prove that Recursive ANNs (R-ANN) with rational weights and ramp activation functions can simulate any $n$-tape ($n\ge 2$) stack machine -- or, equivalently, any Turing machine (TM) and any partial recursive function -- when endowed with a specific localist architecture. Moreover, Siegelmann and Sontag showed that a R-ANN consisting of 886 units can simulate a universal Turing machine (UTM). Recent work by Cabessa \citep{cabessa2012computational, cabessa2012expressive, cabessa2013super} extends these results on R-ANNs to the realm of interactive computation \citep{Wegner98}, a framework studying systems that can interact with the environment throughout their computation (as opposed to the framework of classical computation, where the interaction is limited to the input-output exchange), proving that R-ANNs are equivalent in power to interactive TMs.

Very-large-scale and reservoir-like neural network approaches can also rely on VSA as a key ingredient, as in the \emph{neural engineering} framework \citep{EliasmithEA12, StewartChooEliasmith14}, which employs semantic pointers for addressing symbolic representations in activation space, and recent work at the interface between \emph{reservoir computing} and connectionist/symbolic approaches \citep{HinautDominey13, HinautPetitEA14}

In contrast, the present work focuses on parsimonious VSA implementations, building upon the seminal results from \citet{siegelmann_math_1991, siegelmann_computational_1995}, and work from \citet{Moore90, Moore91a} who has shown that nonlinear dynamical automata (NDA), piecewise-affine linear dynamical systems on the unit square, can simulate the dynamics of any TM in real-time\footnote{In a real-time simulation, a single computation step in the original model is mapped to a single computation step in the model simulating it.} when the machine is represented as a generalized shift (GS) on dotted sequences.
In this work we first extend Moore's results by showing that NDA can support the real-time simulation of a range of models of computation, including but not limited to Turing Machines (of course, TMs can simulate any other model of computation of lesser or equal power, but not necessarily in real-time; see \autoref{sec:VS} for a discussion). We achieve this by relaxing the definition of GS, which leads to a novel and more expressive shift map, the versatile shift (VS) which enables the parsimonious and real-time emulation of symbolic computation in a range of models. We then show that VS dynamics can be mapped to NDA dynamics on the unit square through G\"odelization.
Finally, we present a mapping between VS and R-ANNs through NDA (extending preliminary results shown in \citealp{Carmantini2015Turing}).

Symbolic models of computation distinguish between data, operations on data and the control of these operations. For example, automata implement a set of symbolic operations and its control through a look-up table (the transition function), and the data as a string encoding the so-called \emph{configuration} of the automaton. In grammars and term rewriting systems, operations are instead defined as a set of substitution/rewriting rules on some symbolic string, where the application of these rules is controlled by a set of conditions.
NDA can perform symbolic computation on a vectorial space while preserving, in their formulation, the division between data, operations on data, and their control. Basing our construction on NDA, we derive an architecture that also preserves this division, thus obtaining networks that are transparent, modular and parsimonious. Importantly, the operations embedded within the architecture we propose herein are not only distinguishable in activation space, but are also spatially localized, while still relying on a distributed representation of the symbolic data. The granular modularity of the architecture brought about by its relation with NDA differentiates our approach from previous work, and has important consequences for the constructive mapping of interactive automata networks (IANs) to R-ANNs, and for the possibility of correlational studies with electrophysiological data, which we will discuss in subsequent Sections.

We illustrate our approach by means of two examples.
As a first example, we construct a central pattern generator (CPG) from a finite-state automaton for gait patterns of quadruped animals \citep{GrillnerZangger75, CollinsRichmond94, GolubitskyStewartEA99}.
The neuronal sequential activations by CPGs are usually modeled through networks of coupled nonlinear oscillators that undergo symmetry-breaking bifurcations under changes in their driving input \citep{GolubitskyStewartEA99, GolubitskyStewartEA98, SchonerJiangKelso90, CollinsRichmond94}. We show that our construction, although symbolically inspired, allows the investigation of similar bifurcation scenarios. Additionally, the results of these example are relevant to the design of CPGs for the control of robotic locomotion \citep{ijspeert2008central}.
As a second example, we show how our approach is ideally suited to tackle the mapping of interactive machines to neural networks, because of the separation in the network architecture of data, transformations and their control. This makes it straightforward to construct R-ANNs simulating networks of automata that e.g. share states, are organized in complex hierarchies, or are bound by interactions of conditions in the application of symbolic transformations. We demonstrate this by constructing an interactive automata network (IAN) that implements a diagnosis and repair parser for syntactic language processing \citep{lewis1998reanalysis} and by subsequently mapping it to a R-ANN performing the same computation. We are then able to derive vectorial observables from the network; specifically, we compute synthetic event-related brain potentials (synth-ERPs, \citealp{barres2013synthetic}) and discuss their relation with event-related potentials as measured in experiments involving garden-path sentences~\citep{FrischGrabenSchlesewsky04}.

\begin{table}\centering
  \begin{tabular}{@{}rl@{}}\toprule
    \textbf{Abbreviation} & \textbf{Extended name} \\
    \midrule
    ANN & Artificial neural network \\ 
    BSL & Branch selection layer\\
    CFG & Context-free grammar \\
    CL & Configuration layer\\
    CPG & Central pattern generator\\
    EEG & Electroencephalography\\
    ERP & Event-related brain potentials \\
    FSM & Finite-state machine\\
    GS & Generalized shift \\
    LFP & Local field potentials \\
    LTL & Linear transformation layer\\
    MCL & Machine configuration layer\\
    NDA & Nonlinear dynamical automaton \\
    PCA & Principal component analysis\\
    PDA & Push-down automaton\\
    R-ANN & Recurrent artificial neural network\\
    SRN & Simple recurrent network \\
    synth-ERP & Synthetic event-related brain potential\\
    TDR & Top-down recognizer \\
    TM & Turing machine\\
    UTM & Universal Turing machine\\
    VS & Versatile shift\\
    VSA & Vector symbolic architecture\\
    \bottomrule
  \end{tabular}
  \caption{{\bf List of abbreviations used in this paper.}}
\end{table}

\section{Methods}
\begin{figure}
  \centering
\includegraphics[width=.7\linewidth]{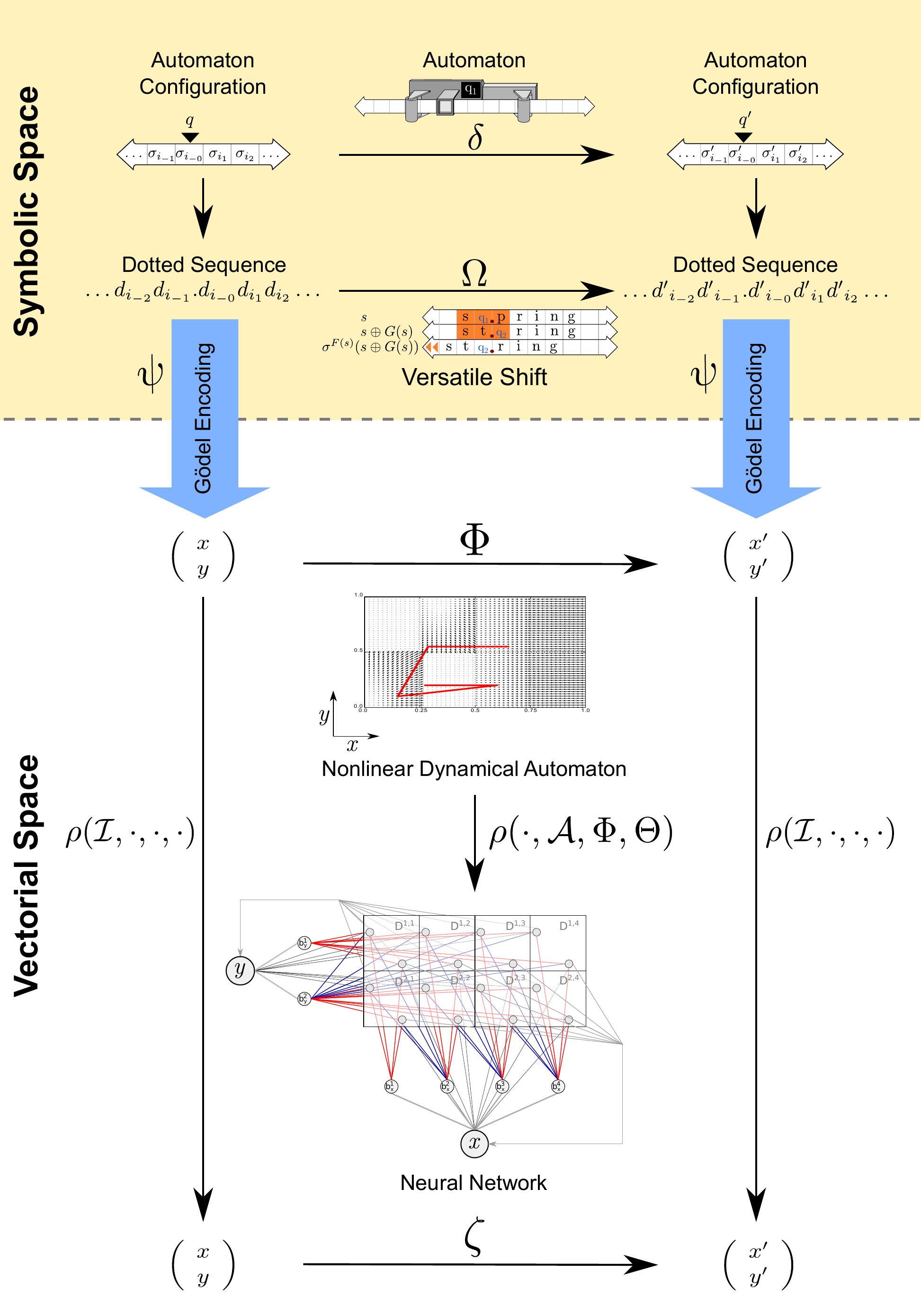}
\caption{{\bf An automaton is mapped to a recurrent artificial neural network (R-ANN).} The representation of machine configurations as dotted sequences allows for the mapping of the machine transition function to the action of a $\Omega$ versatile shift (VS) map upon said sequences, simulating the computation performed by the automaton. A G\"{o}del encoding $\psi$ acts as a bridge between the Symbolic and the Vectorial representation of the automaton's dynamics, and enables the representation of $\Omega$ as an affine-linear map $\Phi$ by a nonlinear dynamical automaton (NDA). Finally, a map $\rho$ generates a R-ANN, with a specific network architecture and internal dynamics $\zeta$ that operates on the same Vectorial space as $\Phi$, where the NDA states are identically mapped through $\rho(I, \cdot, \cdot, \cdot)$ to the activation of a specialized layer in the R-ANN.}
  \label{fig:complete_mapping}
\end{figure}

The present Section outlines our general method which allows the mapping of a range of models of computation to R-ANNs. In~\autoref{fig:complete_mapping} we summarize the complete mapping procedure to accompany its exposition. Our construction is a two-step process.
We first define a Versatile shift (a generalization of the shift map introduced in \citealp{Moore90}) that emulates some model of computation, and we subsequently encode its dynamics on the unit square via G\"{o}delization, obtaining a two-dimensional piecewise affine-linear map on the unit square, i.e. a NDA. 
As a second step, the NDA is mapped onto a first-order R-ANN, which is endowed with an architecture that captures the NDA's three key components: i) a state, encoding the symbolic data of the model of computation; ii) a set of affine-linear transformations, encoding its operations on data; iii) a switching rule that selects the relevant affine-linear transformation to apply given the state, thus implementing the control of the symbolic operations.

Next, the theoretical methods employed are discussed in detail. In the presentation of various objects from Formal Language Theory and Automata Theory, we essentially follow the  well-established definitions in \citet{HopcroftUllman79}, and in \citet{sipser2006introduction}.

\subsection{Elements of Symbolic Computation}

A symbol is meant to be a distinguished element from a finite set $\mathbf{A}$, which we call an \emph{alphabet}. Symbols can be concatenated, i.e. for $a, b \in \mathbf{A}$, $ab \equiv (a, b) \in \mathbf{A}^2$. A sequence of symbols $w \in \mathbf{A}^n$ is called a word of length $n$, denoted $n = |w|$. The set of words of all possible lengths $w$ of finite length $|w| \ge 0$ is denoted $\mathbf{A}^*$ (for $|w| = 0$, $w = \epsilon$ denotes the ``empty word'').

\subsubsection{From Generalized to Versatile Shifts}
\label{sec:VS}

The theory of symbolic dynamics \citep{LindMarcus95} is a tool to study dynamical systems based on the discretization of time and space in order to interpret trajectories in a vectorial space as discrete sequences of infinite strings of symbols. Importantly, its theoretical apparatus can also be used to do the opposite, mapping sequences of strings to a vectorial space. We start by redefining a representation for strings of symbols, the dotted sequence.

According to \citet{Moore90, Moore91a}, a \emph{dotted sequence} $s \in \mathbf{A}^\mathbb{Z}$ on an alphabet $\mathbf{A}$ is a two-sided infinite sequence of symbols ``$s = \ldots \; d_{-2} \; d_{-1} \; . \; d_{0} \; d_{1} \; d_{2} \; \ldots$'' where $d_i \in \mathbf{A}$, for all indices $i \in \mathbb{Z}$. Here, the dot ``.'' is simply used as a mnemonic sign, indicating that the index 0 is to its right. A shift space $M_S = (\mathbf{A}^\mathbb{Z}, \sigma)$ is then given by a shift map $\sigma: \mathbf{A}^\mathbb{Z} \to \mathbf{A}^\mathbb{Z}$ \citep{LindMarcus95}, such that $\sigma(s)_i = (s)_{i+1}$, i.e. $\sigma$ shifts all symbols in $s$ one place to the left (or, equivalently, shifts the dot one place to the right). Similarly, it is possible to define an inverse to the shift map, $\sigma^{-1}$, shifting all symbols in $s$ one place to the right (or, equivalently, the dot one place to the left).

Notice how shifting the dot in a dotted sequence to the left or the right resembles the movement of the read-write head of a Turing machine on its tape (see \autoref{sec:TMs} for more details on Turing machines). In order to fully attain the power of Turing machines, \citet{Moore90, Moore91a} endows the shift space $M_S$ with three additional maps
\begin{equation}
\begin{aligned}
  &F: \mathbf{A}^\mathbb{Z} \rightarrow \mathbb{Z} \\
  &\oplus: \mathbf{A}^\mathbb{Z} \times (\mathbf{A} \cup \{ \phi \})^\mathbb{Z} \rightarrow \mathbf{A}^\mathbb{Z}\\
  &G: \mathbf{A}^\mathbb{Z} \rightarrow (\mathbf{A} \cup \{ \phi \})^\mathbb{Z},
\end{aligned}
\end{equation}
such that their composition $\Omega(s) = \sigma^{F(s)}(s \oplus G(s))$ can fully simulate any Turing machine. The augmented shift space $M_{GS} = (\mathbf{A}^\mathbb{Z}, \Omega)$ is called \emph{generalized shift} (GS) if there is an open interval of indices around the dot, called \emph{Domain of Dependence} $\text{DoD}=(k_l, k_r)$ ($k_l \le 0 \le k_r$), such that $F(s)$ and $G(s)$ only depend on the content of $s$ within the $\text{DoD}$, $F(s)$ determines a number of left shifts ($F(s) >0 $), right shifts ($F(s) < 0$), or no shift at all ($F(s) = 0$) and $G(s)$ maps the symbols $s_i$ within the $\text{DoD}$ onto other symbols $g_i$, while all symbols outside the $\text{DoD}$ are mapped onto an auxiliary symbol $\phi$. Finally, the composition operator overwrites all symbols $s_i$ within the DoD through their images $g_i$ under $G$ while not changing $s$ outside the DoD, i.e. $(s \oplus g)_i = s_i$ if $g_i = \phi$, but $(s \oplus g)_i = g_i$ if $g_i \ne \phi$.\footnote{In his \citeyear{Moore91a} paper, Moore actually defines the DoD of a GS as a finite set of integers which need not be consecutive, and introduces a second finite set of integers, the \emph{Domain of Effect} (DoE) to indicate the cells to be rewritten (as a function of the cells in the DoD). Nevertheless, it is always possible, given any GS with arbitrary DoD and DoE, to construct an equivalent GS as defined here; we thus decided to propose a simplified definition.}

\begin{figure}
  \centering
\includegraphics[width=.7\linewidth]{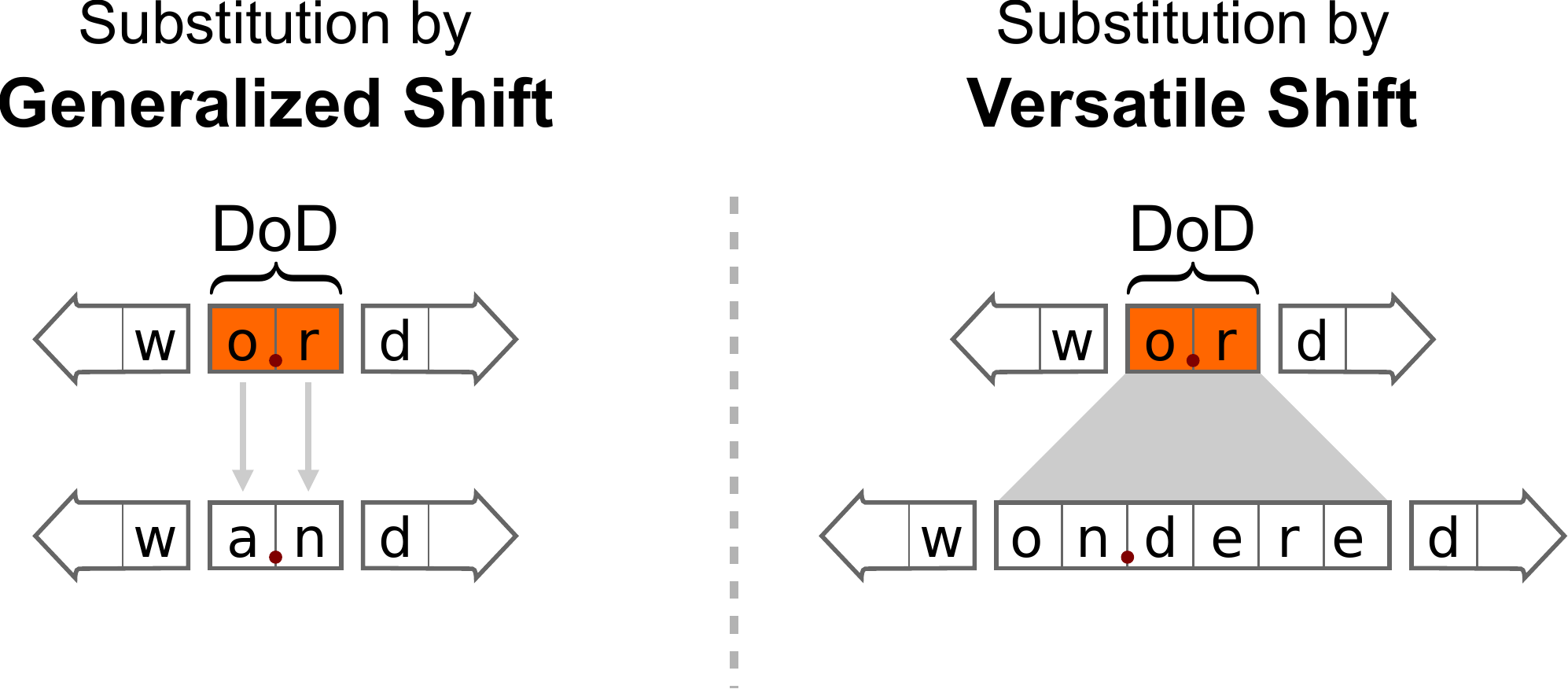}
\caption{{\bf Difference between substitution operation in generalized and versatile shifts.} In this Figure, we show two example substitutions by respectively a generalized shift and a versatile shift. While a generalized shift can only rewrite each symbol of the dotted word in its Domain of Dependence (DoD) with a new one, a versatile shift can substitute the dotted word in its DoD with any other arbitrary dotted word.}
  \label{fig:generalized_vs_versatile}
\end{figure}

According to Moore's proof \citep{Moore90, Moore91a}, any Turing machine can be realized as a GS $M_{GS}$. Since Turing machines can be programmed to simulate the computation carried out by any model of lower or equal computational power, such as finite-state automata or push-down automata, this implies that these can also be described in terms of equivalent GSs. In practice, however, simulating other automata via Turing machines will lead to rather complicated machine tables even for the simplest symbolic algorithms, and thus to unnecessarily complicated shift spaces. In fact, different automata implement different atomic operations, so that a Turing machine can require multiple computation steps to simulate a single computation step of another automaton, even when the automaton is computationally less powerful.
  Therefore, we introduce a novel shift space to which we shall henceforth refer as versatile shift (VS), which will allow us to represent automata configuration dynamics on dotted sequences in a more straightforward and parsimonious fashion, simulating it in real-time.
Our construction essentially relies on a redefinition of the concept of dotted sequence. Above, the dot was only used as a mnemonic symbol without any functional implication. Now, we introduce the dot as a meta-symbol which can be concatenated with two words $v_1, v_2 \in \mathbf{A}^*$ through $v = v_1 . v_2$. Let $\hat{\mathbf{A}}^*$ denote the set of these dotted words. Moreover, let $\mathbb{Z}^{-} = \{i\; |\; i < 0, \; i \in \mathbb{Z} \}$ and $\mathbb{Z}^{+} = \{ i\; | \; i \ge 0, \; i \in \mathbb{Z} \}$ the sets of negative and non-negative indices. We can then reintroduce the notion of a dotted sequence as follows. Let $s \in \mathbf{A}^\mathbb{Z}$ be a bi-infinite sequence of symbols such that $s = w_\alpha v w_\beta$ with $v \in \hat{\mathbf{A}}^*$ as a dotted word $v = v_1 . v_2$ and $w_\alpha v_1 \in \mathbf{A}^{\mathbb{Z}^{-}}$ and $v_2 w_\beta \in \mathbf{A}^{\mathbb{Z}^{+}}$. Through this definition, the indices of $s$ are inherited from the dotted word $v$ and are thus not explicitly prescribed.
Whereas GSs can only rewrite each symbol in their DoD with a new one, VSs are endowed with a more general rewriting operation, substituting dotted words in their DoD with other dotted words of equal or different lengths (as already hinted, yet not implemented, by \citealp{Moore90}). This adds expressiveness to VSs, allowing for the  parsimonious real-time simulation of a range of automata (see \autoref{fig:generalized_vs_versatile} for a pictorial representation of the difference in substitution operations between GSs and VSs).

More formally, we define a VS as a pair $M_{VS} = (\mathbf{A}^\mathbb{Z}, \Omega)$, with $\mathbf{A}^\mathbb{Z}$ being the space of dotted sequences, and $\Omega : \mathbf{A}^\mathbb{Z} \rightarrow \mathbf{A}^\mathbb{Z}$ defined by
\begin{equation}
  \Omega(s) = \sigma^{F(s)}(s \oplus G(s))
\end{equation}
with
\begin{equation}
\begin{aligned}
  &F: \mathbf{A}^\mathbb{Z} \rightarrow \mathbb{Z} \\
  &\oplus: \mathbf{A}^\mathbb{Z} \times \mathbf{A}^\mathbb{Z} \rightarrow \mathbf{A}^\mathbb{Z}\\
  &G: \mathbf{A}^\mathbb{Z} \rightarrow \mathbf{A}^\mathbb{Z},
\end{aligned}
\end{equation}
where the operator ``$\oplus$'' substitutes the dotted word $v_1.v_2 \in \hat{\mathbf{A}}^*$ in $s$ with a new dotted word $\hat{v_1}.\hat{v_2} \in \hat{\mathbf{A}}^*$ specified by $G$, while $F(s) = F|_{\hat{\mathbf{A}}^*}(v_1.v_2)$ determines the number of shift steps as for the GS above.
The action of $F$, $G$ and $\oplus$ in the VS depends on a finite dotted sub-sequence $v_1.v_2$ inside the original dotted sequence $ s = w_\alpha v w_\beta$, as determined by the DoD of the VS, again defined as a set of consecutive integers denoting cell positions on the original dotted sequence. The DoD of a GS can be specified by an open interval $(k_l, k_r)$ on the integers, with $k_l\le 0$ and $k_r \ge 0$. Additionally, for a $\text{DoD}=(k_l, k_r)$, it is useful to define $\text{DoD}_\alpha = (k_l, 0)$ and $\text{DoD}_\beta = (-1, k_r)$ to denote the left and right part of the complete $\text{DoD}$ on dotted sequences $\alpha.\beta$, with $\text{DoD} = \text{DoD}_\alpha \cup \text{DoD}_\beta$. The set $V$ of dotted words that can appear in the DoD of a VS is a subset of $\hat{\mathbf{A}}^*$, and can be defined as $V = \{v \; | \; v=v_1.v_2 \in \hat{\mathbf{A}}^*, \; |v_1| = |\text{DoD}_\alpha|, \; | v_2 | = | \text{DoD}_\beta |\}$.

To illustrate how VSs act on dotted sequences, consider for example the dotted sequence ``\texttt{wo.rd}'', and define a VS $\Omega_\text{ex}$with
\begin{align*}
  \text{DoD} &= (-2, 1) = \{-1, 0\},\\
  G: &\begin{cases}
    \texttt{o.r} \mapsto \texttt{a.n}\\
    \texttt{a.n} \mapsto \texttt{on.dere}, \\
  \end{cases}\\
  F: &\begin{cases}
    \texttt{o.r} \mapsto 0\\
    \texttt{a.n} \mapsto 1,\\
  \end{cases}
\end{align*}
then, applying $\Omega_\text{ex}$ to ``\texttt{wo.rd}'' once yields
\begin{align*}
  \Omega_\text{ex}(\texttt{w\hl{o.r}d}) & = \sigma^{F(\texttt{w\hl{o.r}d})}\big(\texttt{w\hl{o.r}d} \oplus G(\texttt{w\hl{o.r}d})\big) \\
                                        & = \sigma^{F(\texttt{w\hl{o.r}d})}\big(\texttt{w\hl{o.r}d} \oplus \texttt{a.n}\big) \\
                                        & = \sigma^{F(\texttt{w\hl{o.r}d})}\big(\texttt{wa.nd}\big) \\
                                        & = \sigma^0\big(\texttt{wa.nd}\big) \\
                                        & = \texttt{wa.nd}
                                          \intertext{and applying it again to the resulting ``\texttt{wa.nd}'' dotted sequence yields}
                                          \Omega_\text{ex}(\texttt{w\hl{a.n}d}) & = \sigma^{F(\texttt{w\hl{a.n}d})}\big(\texttt{w\hl{a.n}d} \oplus G(\texttt{w\hl{a.n}d})\big) \\
                                        & = \sigma^{F(\texttt{w\hl{a.n}d})}\big(\texttt{w\hl{a.n}d} \oplus \texttt{on}.\texttt{dere}\big) \\
                                        & = \sigma^{F(\texttt{w\hl{a.n}d})}\big(\texttt{won.dered}\big) \\
                                        & = \sigma^1\big({\texttt{won.dered}}\big) \\
                                        & = \texttt{wo.ndered}
\end{align*}
where the DoD of the input string has been highlighted for clarity (again, contrast this with the pictorial representation given in \autoref{fig:generalized_vs_versatile}).
Note that a VS reduces to a GS in the special case when $G$ always substitutes a dotted sequence with one of the same (finite) length in both the left and the right sub-sequences, as in the previous example where $\texttt{w\hl{o.r}d} \oplus  G(\texttt{w\hl{o.r}d}) = \texttt{w\hl{o.r}d} \oplus\texttt{a.n} = \texttt{wa.nd}$.

A point worth noting is that endowing VS with the rewriting capability extends the GS in the direction of semi-Thue systems (also known as string rewriting systems), a universal model of computation introduced by Axel Thue in 1914 (see chapter 7 of~\citealp{MartinRonElaine94}). These rewriting systems play an important role, for example, in algebraic specifications of abstract data structures, equational programming, program transformation and automated theorem proving, where the conditional and successive application of a finite set of rewrite rules transforms a given symbolic structure.

\subsubsection{Simulation of Various Automata by Versatile Shifts}
We will now discuss how a range of automata can be simulated in real-time by VSs by choosing appropriate dotted sequence representations of machine configurations, and by constructing $F$ and $G$ to reproduce the machine's operations and their conditional application.

\paragraph{Finite-state machines}
The finite-state machine (FSM) model of computation has been introduced by \citeauthor{mcculloch1943logical} in 1943 and is widely used to describe systems in many application fields, ranging from computer science to engineering and biology, to name a few. At every step of a computation a FSM is in one of a finite set of states, and it can change its state as a result of an incoming input signal. More formally, a FSM can be defined as a 5-tuple $M_\text{FSM} = (Q, \mathbf{T}, q_0, F, \delta)$, where $Q$ is a finite set of control states, $\mathbf{T}$ is the input alphabet, $q_0 \in Q$ is the starting state, $F \subseteq Q$ is a set of accept states, and $\delta : Q \times \mathbf{T} \to Q$ is a transition function defined as follows:
\begin{equation}
  \delta: (q_t, d_{0_t}) \mapsto q_{t+1},
  \label{eq:FSM_delta}
\end{equation}
where $q_t, q_{t+1} \in Q$ are states, and $d_{0_t} \in \mathbf{T}$ is an input symbol.
At each computation step, a FSM reads its current state $q_t$, consumes (i.e. reads and discards) its current input symbol $d_t$, and transitions to a new state $q_{t+1} = \delta(q_t, d_t)$ as prescribed by its transition function.
It is possible to encode FSM configurations on dotted sequences as
\begin{equation}
  q_t \; . \; d_{0_t} \; d_{1_t} \; \ldots \; d_{n_t}
\end{equation}
where $q_t$, $d_{0_t}$ and $d_{1_t} \ldots d_{n_t}$ are respectively the state, input symbol, and the rest of the unconsumed input of the FSM at time $t$.
A VS simulating a FSM in real-time can be constructed by defining the Domain of Dependence to be $\text{DoD} = (-2, 1) = \{-1, 0\}$, $F$ to always map to $0$, and $G$ so that for all $q_t \in Q, d_t \in \mathbf{T}$:
\begin{equation}
  G: q_t.d_{0_t} \mapsto q_{t+1}. \epsilon \\
    \label{eq:FSM_GS}
\end{equation}
where $q_{t+1}=\delta(q_t, d_{0_t})$.

\paragraph{Push-down automata and Context-Free Grammars}

A push-down automaton (PDA) is a computing machine that has sequential access to its input and can manipulate a stack memory by popping and pushing symbols on top of it. More formally, a PDA can be defined as a 6-tuple $M_\text{PDA} = (Q, \mathbf{N}, \mathbf{T}, q_0, F, \delta)$, where $Q$ is a finite set of control states, $\mathbf{N}$ is the stack alphabet, $\mathbf{T}$ is the input alphabet, $q_0 \in Q$ is the starting state, $F \subseteq Q$ is a set of accept states, and $\delta$ is a transition function. If $F = \varnothing$, the PDA accepts its input when both the input tape and the stack are empty, and it is thus said to accept by \emph{empty stack}.

A Deterministic PDA is a PDA in which any configuration of the machine defines at most one transition. As the mapping of non-deterministic automata computation to Neural Networks is outside the scope of this work, in what follows we will only discuss Deterministic PDAs. Determinism will thus be implied from this point on. The transition function of a PDA is defined as follows:
\begin{equation}
  \delta: Q \times \mathbf{T}\cup\{\epsilon\} \times \mathbf{N} \rightarrow  Q \times (\mathbf{N} \cup \{\epsilon\}).
\end{equation}
At each computation step, a PDA consumes an input symbol, pushes or pops a symbol on the top of its stack, and changes state as prescribed by its transition function applied to the current state $q_t$, currently read input symbol $d_{0_t}$, and the current top-of-stack symbol $s_{0_t}$. In particular, if $s_{0_t} \ldots s_{m_t}$ is the current content of the stack, transitions of the form
\begin{align*}
  &\delta: (q_t, d_{0_t}, s_{0_t}) \mapsto (q_{t+1}, \epsilon) \\
\intertext{apply a pop operation, such that the new stack content becomes equal to $s_{1_t} \ldots s_{m_t}$. Push operations are instead applied by transitions of the form}
  &\delta: (q_t, d_{0_t}, s_{0_t}) \mapsto (q_{t+1}, s_{0_{t+1}}),\\
\intertext{so that the updated stack contains the symbols $s_{0_{t+1}} s_{0_{t}} \ldots s_{m_t}$. Finally, for transitions of the form}
  &\delta: (q_t, \epsilon, s_{0_{t}}) \mapsto (q_{t+1}, \chi),
\end{align*}
the PDA does not consume any input symbol (i.e. it does not access its input at all), but either pops its top-of-stack, if $\chi=\epsilon$, or pushes symbol $\chi$, if $\chi \in \mathbf{N}$.

PDA configurations can be encoded on dotted sequences as follows:
\begin{equation}
  \underbrace{s_{m_t} \; \ldots \; s_{0_t}}_{s_t} \; q_t \; . \; \underbrace{d_{0_t} \; \ldots \; d_{n_t}}_{d_t}
\end{equation}
where $q_t$, $d_t$ and $s_t$ are respectively the state, the unconsumed input and the content of the stack of the automaton in reversed order at time $t$.

A VS simulating a PDA in real-time can be constructed from the PDA's transition function by defining the Domain of Dependence to be $\text{DoD} = (-3, 1) = \{-2, -1, 0\}$, $F$ to always map to $0$, and $G$ so that, given $\delta: (q_t, \kappa, s_{0_t}) \mapsto (q_{t+1}, \chi$),
\begin{equation}
  G:
  \left\{
    \begin{array}{c@{\;}c@{\;}c@{\;}ccc@{\;}c@{\;}c@{\;}c@{\;}c@{\qquad}c@{}}
      s_{0_t} & q_{t} & . & \kappa & \mapsto & & \epsilon & q_{t+1} & . & \epsilon &\text{if }\chi = \epsilon \\
      s_{0_t} & q_{t} & . & \kappa & \mapsto & s_{0_t}  & \chi & q_{t+1} & . & \epsilon &\text{otherwise.}
    \end{array}\right.
\end{equation}

  PDA recognize the class of languages generated by context-free grammars (CFG). PDA and CGFs are thus equivalent in power. A CFG specifies a language, i.e. a set of strings on some alphabet, by defining how its words can be constructed, moving from a distinguished starting symbol and applying substitution rules until a string of unsubstitutable symbols (terminals) is reached.

A CFG can be formally defined as a $4$-tuple $G_{CF} = (\mathbf{N}, \mathbf{T}, R, \mathtt{S})$, where $\mathbf{N}$ is a set of non-terminal symbols, $\mathbf{T}$ is a set of terminal symbols, $R \subset \mathbf{N} \times (\mathbf{N} \cup \mathbf{T})^*$ a set of substitution rules and $\mathtt{S}$ a distinguished start symbol. In particular, each rule in $R$ can be written as $X \rightarrow w$, with $X \in \mathbf{N}$ and $w \in (\mathbf{N} \cup \mathbf{T})^* $.

For example, let us define a CFG $G_\text{ex}$ with $\mathbf{N}=\{\mathtt{S}\}$, $\mathbf{T}=\{\texttt{(},\texttt{[}, \texttt{)}, \texttt{]}\}$, and $R$ containing the rules
\begin{align*}
  \mathtt{S} \rightarrow &\texttt{(} \mathtt{S} \texttt{)} \\
  \mathtt{S} \rightarrow &\texttt{[} \mathtt{S} \texttt{]} \\
  \mathtt{S} \rightarrow &\hphantom{\texttt{[}}\epsilon.
\end{align*}
Then $G_\text{ex}$ generates the language $\mathcal{L}_\text{ex}$ of balanced round and square brackets. By applying the substitution rules we can in fact derive any string in that language. For illustration purposes, an example derivation would be: $\mathtt{S} \rightarrow \texttt{[}\mathtt{S}\texttt{]} \rightarrow \texttt{[(}\mathtt{S}\texttt{)]} \rightarrow \texttt{[(}\texttt{)]} \in \mathcal{L}_\text{ex}$.
It is always possible to construct, given any CFG, a PDA recognizing its language, and viceversa.

\paragraph{Top-down recognizers}
\label{sec:TDR}

In one of the examples presented later in the text, we will make use of top-down recognizers (TDRs, see \citealp{AhoUllman72}) that can process locally unambiguous non-left-recursive CFGs\footnote{A recursive CFG is a CFG including rules $A \to u A v$ that expand a non-terminal symbol $A$ into a string containing the same non-terminal. A CFG is called left-recursive if such rules appear in the form $A \to A w$. A CFG is locally unambiguous if there are no two rules expanding the same nonterminal.}.
TDRs are a subclass of PDA that can simulate rule expansion to accept languages generated by non-left-recursive CFGs.
Given any CFG $G_{CF}$ that is not left-recursive, it is possible to construct a TDR that can parse strings belonging to the context-free language generated by that grammar. If the input string of a TDR constructed from $G_{CF}$ is in the language generated by that grammar (and thus it can be derived by the grammar), then the TDR will end its computation with an empty stack and input, and is said to accept the string by empty stack. We are specifically interested in TDRs that process locally unambiguous CFGs, which have the additional property of needing only one state to perform their computation.
To construct such a TDR from a locally unambiguous non-left-recursive CFG $G_{CF} = (\mathbf{N}, \mathbf{T}, R, \mathtt{S})$ it is sufficient to define its $\delta$ function in the following way:
\begin{equation}
  \delta: \left\{
  \begin{array}{c@{\;}c@{\;}ccc@{\;}c@{\qquad}l@{}}
    (q_0, &a, &a)  &\mapsto & (q_0, &\epsilon) &\text{for all } a \in \mathbf{T}\\
    (q_0, &\epsilon, &X) &\mapsto & (q_0, &w) &\text{for all } (X \rightarrow w) \in R
  \end{array}\right.
  \label{eq:delta_TDR}
\end{equation}
where $X\in\mathbf{N}$ is a non-terminal, $w\in(\mathbf{N} \cup \mathbf{T})^*$ is a string of terminals and non-terminals, and $q_0$ is the TDR's only state. Note that in the definition above we endow TDRs with the additional capability of pushing strings $w$ on the stack rather than single symbols.

As our TDRs only have one state $q_0$, we can describe their machine configuration without referring to the current state.
It is thus possible to encode TDR configurations on dotted sequences as follows:
\begin{equation}
  \underbrace{s_{m_t} \; \ldots \; s_{0_t}}_{s_t} \; . \; \underbrace{d_{0_t} \; \ldots \; d_{n_t}}_{d_t}
\end{equation}
where $d_t$ and $s_t$ are respectively the unconsumed input and the content of the stack of the automaton in reverse order at time $t$.
Similarly, simpler VSs than those needed to simulate PDAs can be constructed from a TDR's transition function, by defining the Domain of Dependence to be $\text{DoD} = (-2, 1) = \{-1, 0\}$, $F$ to always map to $0$ and $G$ to mirror \autoref{eq:delta_TDR} so that
\begin{equation}
    G: \left\{
    \begin{array}{c@{}c@{}ccc@{}c@{}c}
      a &. &a &\mapsto &\epsilon &. &\epsilon\\
      X &. &a &\mapsto &w &. &\epsilon\\
    \end{array}\right.
\end{equation}
for all $a \in \mathbf{T}$, $(X \rightarrow w) \in \mathbf{R}$.

\paragraph{Turing machines}
\label{sec:TMs}
A Turing machine (TM) is an automaton with read-write random access to a two-sided infinite tape \citep{turing1937computable, sipser2006introduction}. TMs are central to the Theory of Computation, and they are thought to be powerful enough to model any physically realizable computation (with assumptions of unbounded resources). A TM has an in-built tape (doubly-infinite one dimensional memory with one symbol capacity at each memory location) and a finite-state controller endowed with a read-write head that follows the instructions encoded by the transition function. At each step of the computation, given the current state and the current symbol read by the read-write head, the controller determines via a $\delta$ transition function the writing of a symbol on the current memory location, a shift of the read-write head to the memory location to the left ($\mathcal{L}$) or to the right ($\mathcal{R}$) of the current one, and the transition to a new state for the next computation step. Formally, a TM~\citep{turing1937computable} can be defined as a 7-tuple $M_\text{TM} = (Q,\mathbf{N}, \mathbf{T}, q_0, \sqcup, F, \delta)$, where $Q$ is a finite set of control states, $\mathbf{N}$ is a finite set of tape symbols also containing the blank symbol $\sqcup$, $\mathbf{T}\subset\mathbf{N}\setminus\{\sqcup\}$ is the input alphabet, $q_0 \in Q$ is the starting state, $F\subset Q$ is a set of `halting' states reached at the end of the computation and $\delta: Q \times \mathbf{T} \to Q \times \mathbf{T} \times \{\mathcal{L}, \mathcal{R} \}$ is a partial transition function, the so-called machine table, that determines the dynamics of the machine. In particular, $\delta$ is defined as follows:
\begin{equation}
  \delta: (q_t, d_{0_t}) \mapsto (q_{t+1}, d_{0_{t+1}}, m)
   \label{eq:delta}
\end{equation}
where $q_t, q_{t+1} \in Q$ are the state of the machine before and after the transition, $d_{0_t}, d_{0_{t+1}} \in \mathbf{N}$ are respectively the read and rewritten symbol, and $m \in \{\mathcal{L}, \mathcal{R}\}$ denotes the shift of the read-write head to the left or to the right.

At a given computation step, the content of the tape together with the position of the read-write head and the current controller state define a machine configuration. It is possible to encode TM configurations on dotted sequences as follows:
\begin{equation}
  s = \underbrace{\ldots\;d_{{-2}_t}\;d_{{-1}_t}\vphantom{M}}_{l_t} \; q_t \;. \;\underbrace{d_{{0}_t} \; d_{{1}_t} \; d_{{2}_t} \; \ldots \vphantom{M}}_{r_t},
\end{equation}
where $l_t$ describes the part of the tape to the left of the read-write head, $r_t$ describes the part to its right, $q_t$ describes the current state of the machine controller, and the central dot denotes the current position of the read-write head, i.e. $d_{{0}_t}$, the symbol to its right.

A VS simulating a TM in real-time can be constructed from the TM's transition function by defining the Domain of Dependence to be $\text{DoD} = (-3, 1) = \{-2, -1, 0\}$, and $G$ and $F$ so that, given $\delta: (q_t, d_{0_t}) \mapsto (q_{t+1}, \hat{d}_{0_t}, m) $,
\begin{equation}
  \begin{array}{cl}
    G: &
    \left\{
      \begin{array}{c@{\;}c@{\;}c@{\;}ccc@{\;}c@{\;}c@{\;}c@{\qquad}l@{}}
        d_{{-1}_t} & q_{t} & . & d_{0_t} & \mapsto & d_{{-1}_t} & \hat{d}_{0_t} & . & q_{t+1} &\text{if } m = \mathcal{R} \\
        d_{{-1}_t} & q_{t} & . & d_{0_t} & \mapsto & q_{t+1}  & d_{{-1}_t} & . & \hat{d}_{0_t} &\text{if } m = \mathcal{L}
      \end{array}\right.\\[1.5em]
    F: &
         \left\{
      \begin{array}{c@{\;}c@{\;}c@{\;}ccc@{\;}c@{\;}c@{\;}c@{\qquad}l@{}}
        d_{{-1}_t} & q_{t} & . & d_{0_t} & \mapsto & \hphantom{d_{{-1}_t}} \mathllap{-1} & \hphantom{\hat{d}_{0_t}} & \hphantom{.} & \hphantom{q_{t+1}} &\text{if } m = \mathcal{R} \\
        d_{{-1}_t} & q_{t} & . & d_{0_t} & \mapsto & \hphantom{d_{{-1}_t}} \mathllap{+1}  & \hphantom{d_{{-1}_t}} & \hphantom{.} & \hphantom{\hat{d}_{0_t}} &\text{if } m = \mathcal{L}
      \end{array}\right.
  \end{array}
  \label{eq:TM_versatile}
\end{equation}
for all $d_{{-1}_t} \in \mathbf{N}$.

The following example will clarify how the VS defined as above (\autoref{eq:TM_versatile}) can simulate a TM. Consider, for instance, the dotted sequence ``\texttt{w$q_0$.ord}'', and define a TM such that $\delta: (q_0, \texttt{o}) \mapsto (q_1, \texttt{a}, \mathcal{R})$ and $\delta: (q_1, \texttt{r}) \mapsto (q_1, \texttt{n}, \mathcal{L})$. Then a computation step of the TM starting from the ``\texttt{w$q_0$.ord}'' configuration would yield a new configuration ``\texttt{wa$q_1$.rd}''; by running the TM again, this time starting from ``\texttt{wa$q_1$.rd}'', a computation step would yield ``\texttt{w$q_1$.and}'', as prescribed by the transition function we defined.
Constructing a VS $\Omega_\text{ex}$ as specified by \autoref{eq:TM_versatile} and applying it to ``\texttt{w$q_0$.ord}'':
\begin{equation}
  \begin{aligned}
    \Omega_\text{ex}(\texttt{\hl{w$q_0$.o}rd}) & = \sigma^{F(\texttt{\hl{w$q_0$.o}rd})}\big(\texttt{\hl{w$q_0$.o}rd} \oplus G(\texttt{\hl{w$q_0$.o}rd})\big) \\
    & = \sigma^{-1}\big(\texttt{\hl{w$q_0$.o}rd} \oplus \text{\texttt{wa.$q_1$}} \big) \\
    & = \sigma^{-1}\big(\texttt{wa.$q_1$rd}\big) \\
    & = \texttt{wa$q_1$.rd}
    \label{eq:ex_left_shift}
  \end{aligned}
\end{equation}
and by applying it again to the resulting ``\texttt{wa$q_1$.rd}'' dotted sequence we obtain
\begin{equation}
  \begin{aligned}
    \Omega_\text{ex}(\texttt{w\hl{a$q_1$.r}d}) & = \sigma^{F(\texttt{w\hl{a$q_1$.r}d})}\big(\texttt{w\hl{a$q_1$.r}d}\oplus G\big(\texttt{w\hl{a$q_1$.r}d}\big) \\
    & = \sigma^{+1}\big(\texttt{w\hl{a$q_1$.r}d} \oplus \texttt{$q_0$a.n}\big) \\
    & = \sigma^{+1}\big(\texttt{w$q_0$a.nd}\big) \\
    & = \texttt{w$q_0$.and}
    \label{eq:ex_right_shift}
  \end{aligned}
\end{equation}
where the DoD of the input string to the VS has been highlighted for clarity.
Note that the dotted representation of the machine configuration requires index $-1$ to always contain the machine state. For this reason, it is not enough to only rewrite the symbols in $\{-1, 0\}$ (i.e. the machine state and the current symbol under the read-write head) to simulate a TM, as intuition would instead suggest. In fact, a VS first applies a rewriting of its DoD, and then shifts the resulting dotted sequence to the left (when $F(s)=-1$) or the right (when $F(s)=+1$). In particular, the shift is needed to simulate the movement of the read-write head on the machine tape. In order to make sure that at the end of the substitution and shift the machine state is correctly placed at its reserved index $-1$, the substitution must leave it displaced one place to the right if a left shift is to be applied (as in \autoref{eq:ex_left_shift}), or one to the left in case of a right shift (as in \autoref{eq:ex_right_shift}). This last case requires the additional dependence of the VS on index $-2$. Furthermore, note that our construction is equivalent to that from \citet{Moore90, Moore91a}: the VS defined in \autoref{eq:TM_versatile} is nothing more than the GS introduced by Moore to prove the equivalence between GSs and TMs.

\subsection{Introducing Nonlinear Dynamical Automata}
We will now discuss how VSs, and thus the models of symbolic computation they can simulate, can be mapped to piecewise affine-linear systems on a vectorial space, obtaining nonlinear dynamical automata.

\subsubsection{G\"{o}del Encodings and the Symbol Plane}
\label{sec:symbologram}

\begin{figure}
  \centering
\includegraphics[width=.7\linewidth]{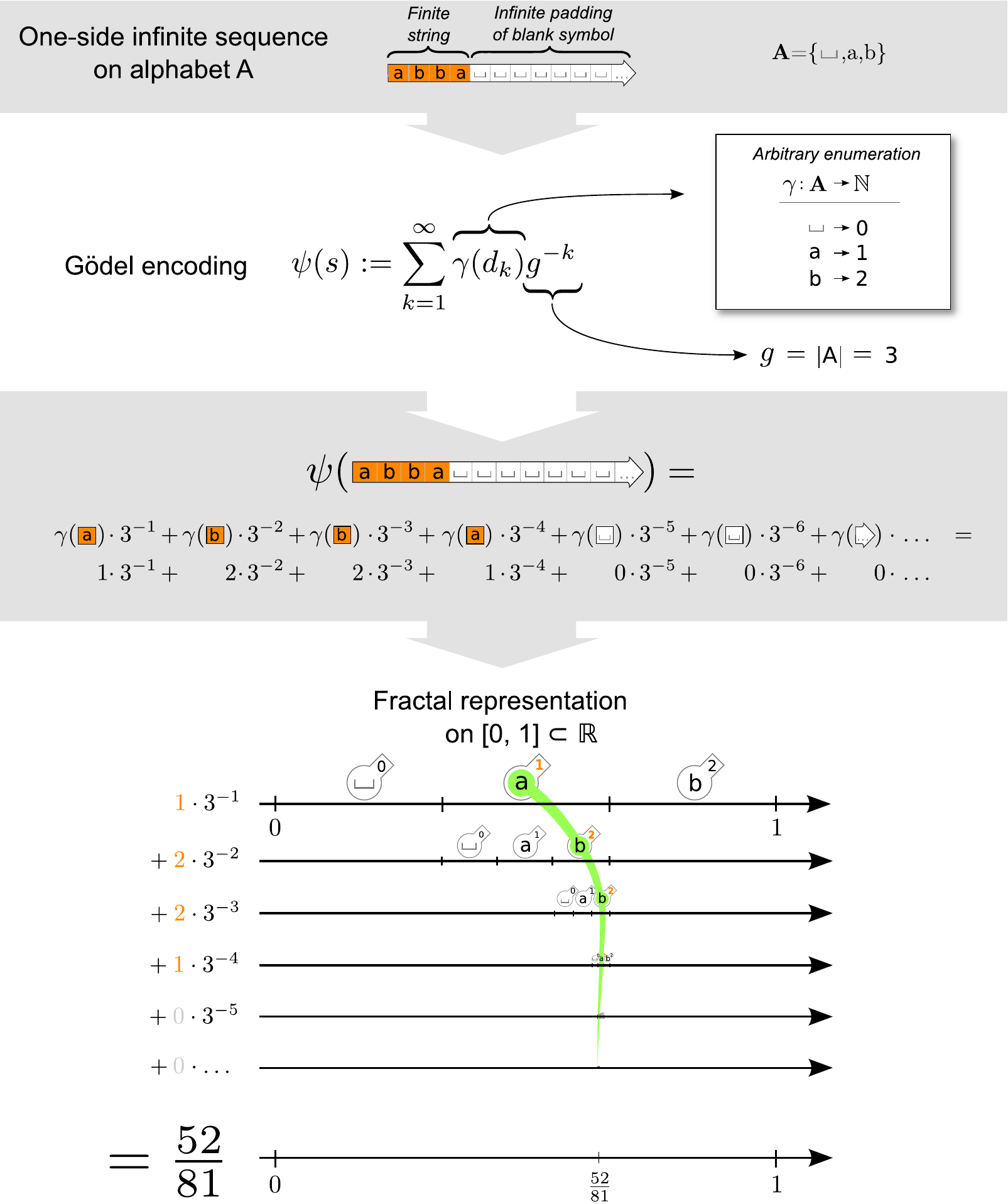}
\caption{{\bf Three representations of the G\"{o}del Encoding of a sequence.} The first one is just the definition of the G\"{o}del encoding, with details on the specific choice of the enumerating $\gamma$ function and the induction of the $g$ constant, given the alphabet $\mathbf{A}$ from which the sequence takes its symbols. The second one is an expansion of the series in the definition. The third one visually conveys the fractal and convergent nature of the series, highlighting the relation between numbers and symbols by the use of the color orange. At each level of this representation, from top to bottom, the encoding of the sequence ``$\mathtt{abba\sqcup\sqcup\sqcup\ldots}$'' is sequentially constructed, highlighting the contribution of each encoded symbol to the real number resulting from the complete G\"{o}delization.}
  \label{fig:ge_of_sequence}
\end{figure}

A G\"{o}del encoding (or G\"{o}delization, see \citealp{Goedel31}) allows one to uniquely assign a real number to a sequence such that the space of one-sided infinite sequences can be mapped to the real interval $[0, 1]$.\footnote{A G\"{o}del encoding maps sequences on some alphabet $\mathbf{A}$ to real numbers through the use of a base-$b$ expansion, with $b=|\mathbf{A}|$. It can be proven that any base-$b$ expansion represents a real number, and that any real number has a unique base-$b$ representation under a weak condition. The uniqueness of the G\"{o}del encoding (and decoding) of any sequence follows from the same proof.} For completeness, G\"{o}delization is subsequently discussed alongside its graphical representation, provided in~\autoref{fig:ge_of_sequence}.\\
Let $\mathbf{A}^\mathbb{N}$ be the space of one-sided infinite sequences over an alphabet $\mathbf{A}$ containing $\vert\mathbf{A}\vert = g$ symbols, and $s=d_1 d_2 \ldots$ a sequence in this space, with $d_k$ being the $k$-th symbol in $s$. Additionally, let $\gamma: \mathbf{A}\rightarrow\mathbb{N}$ be a one-to-one function associating each symbol in the alphabet $\mathbf{A}$ with a natural number. Then a G\"{o}delization is a mapping from $\mathbf{A}^\mathbb{N}$ to $[0,1]\subset \mathbb{R}$ defined as follows:
\begin{equation}
  \psi(s) := \sum\limits_{k=1}^{\infty} \gamma(d_k)g^{-k}.
  \label{eq:godel_encoding}
\end{equation}
Conveniently, G\"{o}delization can also be employed on a dotted sequence $\alpha.\beta \in \mathbf{A}^\mathbb{Z}$ --- herein representing a machine configuration --- by splitting it into its two one-sided constituents $\alpha'$ (the reversed $\alpha$) and $\beta$. Defining two G\"{o}del encodings $\psi_{x}$ and $\psi_{y}$ for $\alpha'$ and $\beta$ respectively, induces a two-dimensional representation for $\alpha.\beta$, i.e. $\big(\psi_{x}(\alpha'), \psi_{y}(\beta)\big)$, known as symbol plane or symbologram, which is contained in the unit square ${[0,1]}^2 \subset \mathbb{R}^2$.
In encoding dotted sequences $\alpha.\beta$ representing configurations of the machines we consider in this paper, $\alpha$ often only ever contains states as first symbols, and tape symbols in the rest of the sequence. In this case we can define a more refined G\"{o}delization that covers all of the representational space $[0,1] \in \mathbb{R}$:
\begin{equation}
    \psi_x(s) := \gamma_q(d_1)n_q^{-1} +
    \sum\limits_{k=1}^{\infty} \gamma_s(d_{k+1})n_s^{-k}n_q^{-1},
  \label{eq:refined_encoding}
\end{equation}
where $\gamma_q$ and $\gamma_s$ respectively enumerate the set of states $Q$ and the tape alphabet $\mathbf{A}$, and where $n_q = \vert Q \vert$, $n_s = \vert \mathbf{A} \vert$.

\subsubsection{Versatile Shifts as Affine-Linear Transformations}
\label{sec:lintransf}

Push and pop operators can be defined on one-sided infinite sequences $\mathbf{A}^{\mathbb{N}}$ on some alphabet $\mathbf{A}$. The push operator $\opush$ is defined so that $s \opush b$ adds the contents of a word $b \in \mathbf{A}^*$ to the beginning of $s \in \mathbf{A}^{\mathbb{N}}$, whereas the pop operator $\opop$ is defined so that $\opop^p s$ removes the first $p$ symbols in $s$. We will now show that G\"{o}delizing a sequence resulting from the application of pop and push operations is equivalent to applying an affine-linear transformation on the original G\"{o}delized sequence. We will then show that VSs on a dotted sequence $\alpha.\beta$ can be mapped to push and pop operations on its one-sided constituents $\alpha'$ and $\beta$. Let $s = d_1 d_2 d_3 \ldots$ be a one-sided infinite sequence on an alphabet $A$. Applying a pop operation $\opop^{p}$ to $s$ yields $\opop^{p}s = d_{p+1} d_{p+2} d_{p+3} \ldots$, while pushing a word $b = b_1 \ldots b_r$ to the beginning of $s$ yields $s \opush b = b_1 \ldots b_r d_{1} d_{2} \ldots$\ . In this case
\begin{align*}
  \psi\left(s\right)& = \gamma(d_{1})g^{-1} + \gamma(d_{2})g^{-2} +
                      \gamma(d_{3})g^{-3} + \ldots, \\
  \shortintertext{so that}\\
  \psi(\opop^p s)& = \gamma(d_{p+1})g^{-1} + \gamma(d_{p+2})g^{-2} + \gamma(d_{p+3})g^{-3} + \ldots\\
                    & = \psi(s) \cdot g^p - \sum\limits_{i=1}^p \gamma(d_i)g^{p-i},\\
  \shortintertext{and}\\
  \psi(s \opush b)  & = \gamma(b_1)g^{-1} + \ldots + \gamma(b_r)g^{-r} +\\
                    & \hphantom{{}={}} \gamma(d_1)g^{-(r+1)} + \gamma(d_2)g^{-(r+2)} + \ldots\\
                    & = \psi(s) \cdot g^{-r} + \sum\limits_{i=1}^r\gamma(b_i)g^{-i},
\end{align*}
proving that the resulting G\"{o}delized sequences can be obtained by applying affine-linear transformations to the original G\"{o}delized sequences. For both pop and push operations, the parameters of the affine-linear transformations only depend on the number and identities on the symbols that are respectively removed from or added to the beginning of the original sequence.
This is of particular importance in the framework of interactive computation \citep{Wegner98}, where the newly added symbol stems from the network's interaction with its environment. Accordingly, the symbol $b$ becomes represented by a linear operator acting on the system's state space, analogous to quantum operators acting on Hilbert spaces \citep{graben2008towards}.

As previously discussed, a VS defines two operations on dotted sequences, a substitution operation $s \oplus G(s)$ which replaces the dotted sub-sequence in the DoD of the shift with a new dotted sequence $G(s)$, and a shift operation $\sigma^{F(s)}$ shifting the symbols in $s$ to the left or to the right by $F(s)$ positions. Let $s \oplus G(s) = w_\alpha u.v w_\beta \oplus \hat{u}.\hat{v}$ be a substitution replacing the dotted sub-sequence $u.v$ in $s$ with the dotted word $\hat{u}.\hat{v}$, then $s \oplus G(s)$ can be straightforwardly mapped to pop and push operations on $u'{w_\alpha}'$ and $v w_\beta$, the one-sided constituents of the original dotted sequence $s$, as follows:
\begin{align*}
  w_\alpha u.v w_\beta \oplus \hat{u}.\hat{v} & = \big((\opop^{\vert u' \vert} u' {w_\alpha}') \opush \hat{u}' \big)' . \big((\opop^{\vert v \vert} v w_\beta) \opush \hat{v}\big) \\
                                    & = \left({w_\alpha}' \opush \hat{u}' \right)' . \left(w_\beta \opush \hat{v} \right) \\
                                    & = w_\alpha \hat{u}. \hat{v} w_\beta\,
\end{align*}
showing that substitutions on dotted sequences can be mapped to pop and push operations on its one-sided constituents.
A left shift $\sigma^{-1}$ and a right shift $\sigma^{1}$ on a dotted sequence $\alpha.\beta = \ldots \; d_{-2} \; d_{-1} \; . \; d_{0} \; d_{1} \; \ldots$ can be mapped to push and pop operations on its one-sided constituents as follows:
\begin{align*}
  \sigma^{-1}(\ldots \; d_{-2} \; d_{-1} \; . \; d_{0} \; d_{1} \; \ldots) & = (\alpha' \opush d_{0})'. (\opop^1 \beta) \\
                                                                           & = \ldots \; d_{-1} \; d_{0} \; . \; d_{1} \; d_{2} \; \ldots\ , \\
    \shortintertext{and}
  \sigma^{1}(\ldots \; d_{-2} \; d_{-1} \; . \; d_{0} \; d_{1} \; \ldots) & = (\opop^1 \alpha')'. ( \beta \opush d_{-1}) \\
                                                                           & = \ldots d_{-3} \; d_{-2} \; . \; d_{-1} \; d_{0} \; \ldots\ , \\
\end{align*}
showing that shifts on dotted sequences can be mapped to pop and push operations on its one-sided constituents.
Any arbitrary shift $\sigma^k$ with $k\in \mathbb{Z}$ can be obtained by composition of left and right shifts; as the composition of affine-linear transformations is an affine-linear transformation, the G\"{o}delization of a sequence resulting from the composition of shift operations is equivalent to an affine-linear transformation on the original G\"{o}delized sequence. We have thus shown that VSs on dotted sequences can be mapped to pop and push operations on one-sided infinite sequences, and that the G\"{o}delization of these operations can be mapped to affine-linear transformations on the original sequences.
On the symbologram, each substitution and shift operation on a G\"{o}delized dotted sequence $\alpha.\beta$ by a VS involves two affine-linear transformations, one acting on the G\"{o}delized $\alpha'$ (the reversed $\alpha$) and one on the G\"{o}delized $\beta$. The parameters of the affine-linear transformations only depend on the symbols of the dotted sequence in the DoD of the VS. All dotted sequences which share the same DoD symbols are thus associated to the same pair of affine-linear transformations. For this reason, the symbologram representation of VSs leads to piecewise affine-linear maps on rectangular partitions of the unit square, referred to as a nonlinear dynamical automata \citep{Tabor00a, TaborChoSzkudlarek13, graben2004language, graben2008towards}.

\subsubsection{Nonlinear Dynamical Automata}
\label{sec:NDA}
A nonlinear dynamical automaton (NDA) is a triple $M_{NDA} = (X, P, \Phi)$, where $P$ is a rectangular partition of the unit square $X={[0, 1]}^2 \subset \mathbb{R}^2$, that is
\begin{equation}
  P = \{D^{i,j} \subset X |~ 1 \le i \le m,\; 1 \le j \le n,\; \space m,n \in \mathbb{N}\},
  \label{Eq:partition}
\end{equation}
so that each cell is defined as $D^{i,j} = I_i \times J_j$, with $I_i, J_j \subset [0,1]$ being real intervals for each bi-index $(i,j)$, with $D^{i,j} \cap D^{k,l} = \varnothing$ if $(i,j) \neq (k,l)$, and $\bigcup_{i,j}D^{i,j}=X$. The couple $(X, \Phi)$ is a time-discrete dynamical system with phase space $X$ and the flow $\Phi: X \rightarrow X$ is a piecewise affine-linear map such that $\Phi_{|D^{i,j}}:=\Phi^{i,j}$, with $\Phi^{i,j}$ having the following form:

\begin{equation}
  \Phi^{i,j}(\mathbf{x}) = \left(\begin{array}{c} a^{i,j}_x\\ a^{i,j}_y \end{array}\right) +
  \left(\begin{array}{cc} \lambda^{i,j}_x & 0 \\ 0 & \lambda^{i,j}_y \end{array}\right)
  \left(\begin{array}{c} x\\ y \end{array}\right).
  \label{Eq:NDA_dynamics}
\end{equation}
Note that the NDA, as any piecewise affine-linear system, also requires a switching rule $\Theta(x,y) \in \{(i,j)|\;1\leq i\leq m, 1\leq j\leq n\}$, which selects the appropriate branch, and thus dynamics (i.e. $\Phi(x,y) = \Phi^{i,j}(x, y) \iff \Theta(x,y) = (i,j)$). A mapping between a VS and a NDA can be defined following the methods outlined in~\autoref{sec:symbologram} and~\autoref{sec:lintransf}, therefore enabling the derivation of the parameters of the NDA. That is, first each cell $D^{i,j} = I_i \times J_j$ can be seen as containing all the G\"{o}delized dotted sequences $\alpha.\beta$ which agree (i.e.~have the same symbols) in the Domain of Dependence. In particular, the $I_i$ interval contains all the DoD-agreeing G\"{o}delized $\alpha'$ (the reversed $\alpha$) sub-sequences, whereas the $J_j$ interval contains all the DoD-agreeing G\"{o}delized $\beta$ sub-sequences. This leads to a partition of the unit square with a number $i$ of $I$ intervals equal to the number of possible one-sided sub-sequences that can appear in the left DoD of the VS, and a number $j$ of $J$ intervals equal to the number of possible one-sided sub-sequences that can appear in the right DoD. For example, for a VS simulating a FSM, the left Domain of Dependence $\text{DoD}_\alpha=\{-1\}$ of the dotted sequences representing machine configurations only ever contains states, and the right Domain of Dependence $\text{DoD}_\beta=\{0\}$ only ever contains input symbols. In this case the number of $I_i$ intervals becomes equal to the number of states $n_q = \vert Q \vert$ in the FSM, and the number of $J_j$ intervals equal to the number of input symbols $n_s = \vert \mathbf{T} \vert$, where $Q$ and $\mathbf{T}$ are respectively the set of states and that of input symbols in the FSM. For a VS simulating a TM, instead, the left Domain of Dependence $\text{DoD}_\alpha=\{-2, -1\}$ only ever contains states at index $-1$, and tape symbols at index $-2$, and the right Domain of Dependence $\text{DoD}_\beta=\{0\}$ always contains tape symbols. This leads to a partition of the unit square with a number of $I_i$ intervals equal to $m = n_q n_s$, and one of $J_j$ intervals equal to $ n = n_s$, leading to a total of $n_q n_s^2$ cells, where $n_s$ is the number of symbols in the tape alphabet $\mathbf{N}$ and $n_q$ is the number of states in $Q$.

Following \autoref{sec:lintransf}, substitutions and shifts on a sequence can be mapped to affine-linear transformations on its G\"{o}delization. For this reason, each cell in the partition $P$ of the unit square is associated with a different affine-linear transformation with parameters $(a^{i,j}_x, a^{i,j}_y)$ and $(\lambda^{i,j}_x, \lambda^{i,j}_y)$, which can be derived using the methods outlined in \autoref{sec:lintransf}. Therefore a model of computation can be represented as a NDA by means of its G\"{o}delized VS representation.

\subsection{Solution Map between NDA and R-ANNs}
\label{sec:NDAtoRANN}
The design of the map between the NDA and a first order R-ANN follows a conceptually natural and simple solution, which attempts to mimic the affine-linear dynamics (given by~\autoref{Eq:NDA_dynamics}) of the NDA on the partitioned unit square (see \citealp{Carmantini2015Turing} for preliminary work in this direction).

\begin{figure}
  \centering
\includegraphics[width=.35\linewidth]{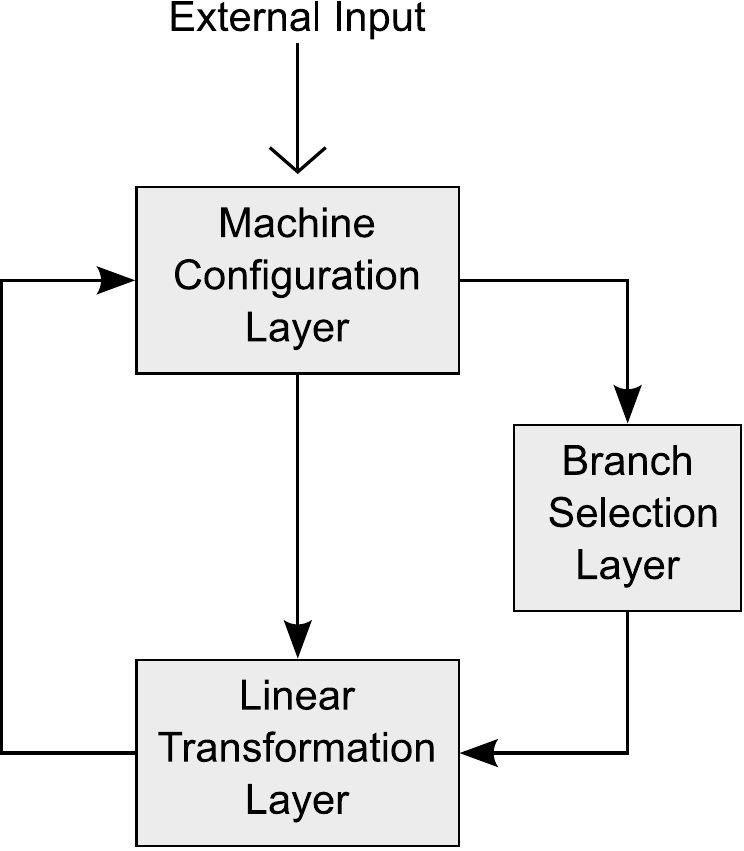}
\caption{ {\bf Connectivity between neural layers within the network.} The machine configuration layer (MCL) receives external input (in this case the encoded initial machine configuration), and synaptically couples to the branch selection layer (BSL) and linear transformation layer (LTL). The BSL feed-forwards to the LTL and finally the LTL recurrently feedbacks to the MCL, where the output is read-out.}
  \label{fig:network_architecture}
  \end{figure}

Let $\rho(\cdot)$ denote the proposed map. The objective is to map the orbits of the NDA (i.e. $\Phi^{i,j}(x,y)$) to orbits of the R-ANN, denoted as $\zeta^{i,j}(x,y)$. The role of $\rho$ is to encode both the affine-linear dynamics within each partition cell ($D^{i,j}$) and to emulate the transitions from cell to cell by suitably activating certain neural units within the R-ANN. To achieve this, we propose a network architecture with three layers, namely a machine configuration layer (MCL), a branch selection layer (BSL) and a linear transformation layer (LTL), as depicted in~\autoref{fig:network_architecture}. Therefore, we generically define the proposed map as follows:
\begin{equation}
  \zeta = \rho(\mathcal{I}, \mathcal{A}, \Phi, \Theta),
\end{equation}
where $\mathcal{I}_{2 \times 2}$ is the identity matrix that maps (identically) the initial conditions of the NDA to the R-ANN and $\mathcal{A}$ is the synaptic weight matrix that defines the network architecture, which will be discussed in subsequent Sections. In addition, $\rho$ generates different neural dynamics for each type of the neural units, i.e. $\zeta=(\zeta_1,\zeta_2,\zeta_3)$, corresponding to MCL, BSL and LTL, respectively. The details of the R-ANN architecture and its dynamics will now be presented.

\subsubsection{Network Architecture and Neural Dynamics}
\begin{figure}
  \centering
\includegraphics[width=\linewidth]{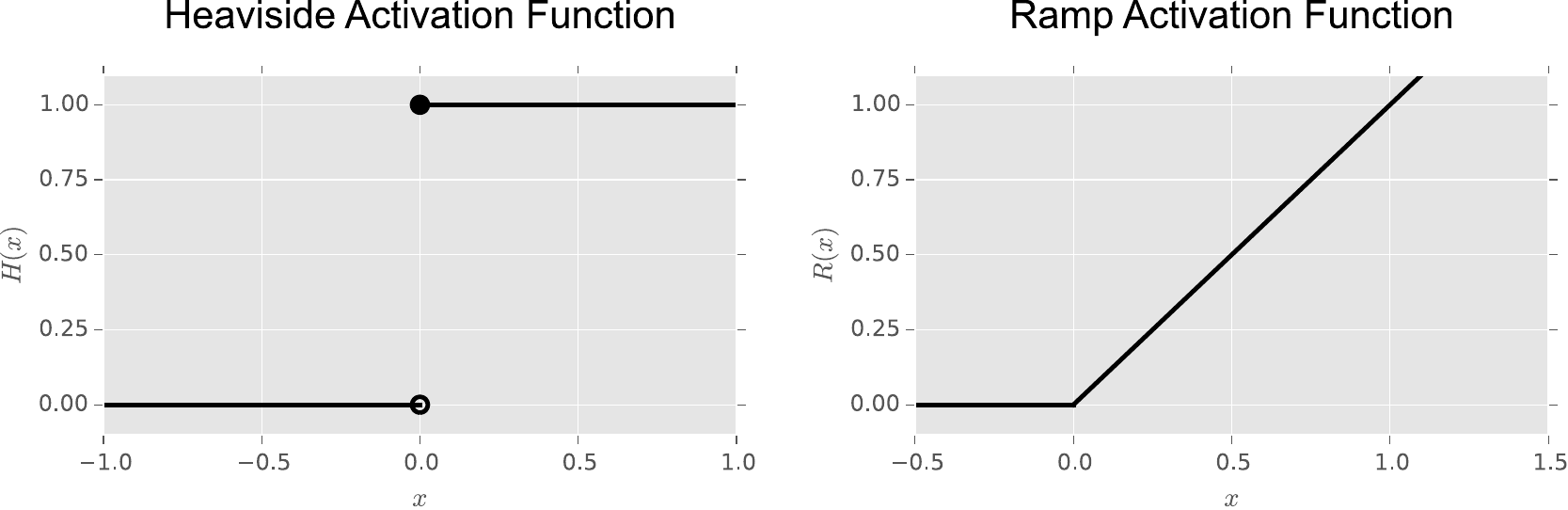}
\caption{{\bf Activation functions employed in the network.} In particular, the Heaviside function $H(x)$ is employed by units in the branch selection layer and the Ramp function $R(x)$ is used in both machine configuration layer and linear transformation layer.}
  \label{fig:act_functions}
\end{figure}

The simulation of a NDA orbit within the R-ANNs is distributed among MCL, BSL and LTL. Since $\Phi^{i,j}(x)$ is a two-dimensional de-coupled discrete map it suggests only two neural units in a read-out layer, which is a role taken by the MCL. We refer to the two MCL units as $c_x$ and $c_y$. At each computation step the MCL stores the encoding of the current machine configuration, which is then passed on to the BSL and LTL units. Subsequently, two sets of BSL units ($b_x$ and $b_y$) functionally act as a switching  system that determines to which cell $D^{i,j}$ the current machine configuration belongs, triggering the appropriate units within two sets of LTL units ($t_x$ and $t_y$), effectively emulating the application of an affine-linear transformation $\Phi^{i,j}$ on an encoded machine configuration. This action corresponds to the application of a symbolic operation by the original machine, leading to a configuration update. The result of the transformation is then fed back to the MCL, representing the configuration (i.e. the machine's symbolic data) for the next computation step. These successive transformations effectively emulate the action of a NDA, where for every computational step an affine-linear transformation is applied to the values encoding the representation of the machine configuration. 
The neural units in the various layers make use of either the Heaviside ($H$) or the Ramp ($R$) activation functions defined as follows (see also \autoref{fig:act_functions}):

\begin{align}
    H(x) &=
    \begin{cases}
      0 &\mbox{if } x<0\\
      1 &\mbox{if } x\ge0\\
    \end{cases} &
    R(x) & =
    \begin{cases}
      0 &\mbox{if\;} x < 0\\
      x &\mbox{if\;} x \ge 0.\\
    \end{cases}
    \label{eq:activation_functions}
\end{align}

\subsubsection{Machine Configuration Layer}
\begin{figure}
  \centering
\includegraphics[width=\linewidth]{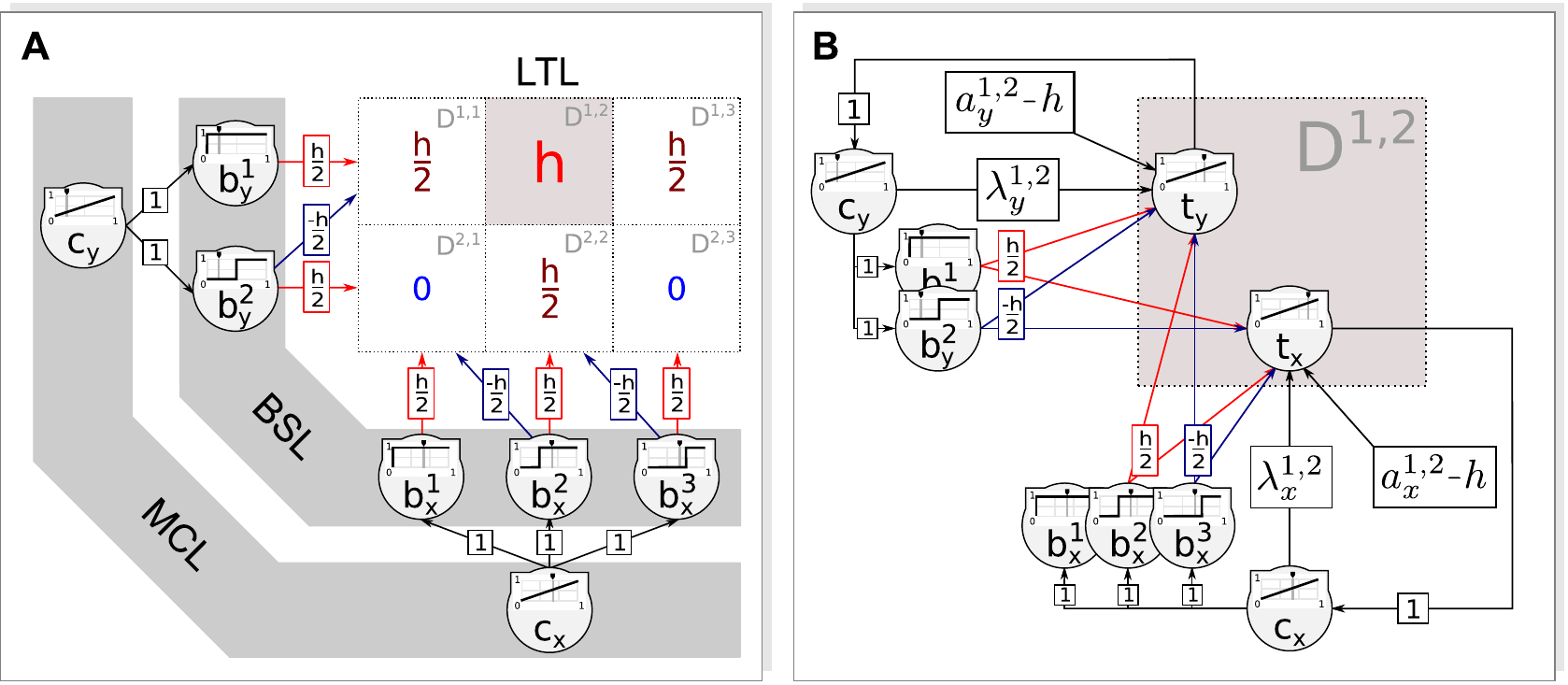}
\caption{{\bf Detailed feedforward connectivity and weights for neural network simulating a nonlinear dynamical automaton with only 6 branches.} (A) The machine configuration layer (MCL) units $(c_x, c_y)$ feed-forward connect to all the branch selection layer (BSL) units with weight of $1$. Every BSL unit excites with weights $\frac{h}{2}$ (in red) and also inhibits with weights $\frac{h}{2}$ (in blue) the relevant linear transformation layer (LTL) units contained within each cell (as indicated by the red and blue arrows respectively). Each cell $D^{ij}$ indicates the overall summed input value received by each LTL unit (for visualization purpose/convenience not shown) from the BSL. In this case only the LTL units in cell $D^{1,2}$ are activated with overall BSL input value of $h$ (red). (B) A zoom-in of panel (A), shows in detail how each pair of LTL units contained within each cell (in this case $D^{1,2}$) receives inputs from the MCL and BSL units as shown. In addition, the LTL units may have internal dynamics described by parameter $a$ (equivalently, this can be seen as input from an always-active unit). To actually produce output, the overall input to an LTL unit must overcome its internal $h$ inhibition. Upon activation, the LTL unit's output is fed back to the paired MCL unit with weight of 1.}
  \label{fig:full_BSL}
\end{figure}

The MCL encodes the state of the simulated NDA, and thus the data of the simulated automaton, while acting as a read-out neural layer. At the same time it mediates at each computation step the transmission of the current G\"{o}del encoding of the emulated machine's configuration to the BSL and LTL units. Since the G\"{o}del encoding of a dotted sequence representing a machine configuration consists of two values (see~\autoref{sec:symbologram}), this implies that the MCL solely requires two neural units ($c_x$ and $c_y$) to code for the current configuration. As a consequence, the initialization of the R-ANNs is performed in this layer, where the initial conditions $(\psi_x(\alpha'), \psi_x(\beta))$ are identically transformed (via $\mathcal{I}$) by the map $\rho(\cdot)$ as follows:
\begin{equation}
  (c_x, c_y) =  (\psi_x(\alpha'), \psi_x(\beta)) \equiv \zeta_1 = \rho(\mathcal{I}, \cdot , \cdot, \cdot)_{| (\psi_x(\alpha'), \psi_x(\beta))}
\end{equation}
Following every computation step, these neural units receive inputs from the LTL units and are subsequently activated via the ramp activation function (\autoref{eq:activation_functions}); in other words $\zeta_1 \equiv (c_x, c_y )= (R(\sum_i t^{i}_x), R(\sum_j t^{j}_y)$). Finally, these synaptically project onto the BSL and LTL neural units (refer to \autoref{fig:full_BSL} for details of the connectivity).

\subsubsection{Branch Selection Layer}
\label{sec:BSL}
The BSL acts as a control unit that enables the sequential mapping of the orbits of the NDA, $\Phi^{i,j}(x,y)$, to orbits of the R-ANNs, $\zeta^{i,j}(x,y)$. Specifically, the BSL functionally embodies the switching rule $\Theta(x,y)$ and coordinates the dynamic switching between LTL units. Sequentially, under the action of BSL units, only a single pair of LTL units $(t^{i,j}_x, t^{i,j}_y)$ dedicated to emulate $\Phi^{i,j}$ become active, which then operate on an encoded Machine configuration. In particular, the BSL units make sure that $(t^{i,j}_x, t^{i,j}_y)$ become active only if $(c_x, c_y) \in D^{i,j} = I_i \times J_j$, with $I_i = [\xi_i, \xi_{i + 1})$ being the $i$-th interval on the $x$-axis and $J_j = [\eta_j, \eta_{j + 1})$ being the $j$-th interval on the $y$-axis. The switching rule is mapped by $\rho(\cdot)$ as follows:
\begin{eqnarray}
  \zeta_{2}(x,y) = \rho(\cdot, \cdot, \cdot, \Theta(x,y)=\{i,j\})
\end{eqnarray}
The implementation of $\zeta_{2}(x,y)$ is mediated by two sets of neural units, i) the $b_x$ set with $m$ units (the number of $I$ intervals on the $x$-axis) and ii) the $b_y$ set with $n$ units (the number of $J$ intervals on the y axis), which are activated via a Heaviside activation function (\autoref{eq:activation_functions}) after receiving excitatory inputs with synaptic weight $1$ from the MCL layer (i.e. $c_x$ and $c_y$ units) in the following way:
\begin{equation}
  \begin{aligned}
    b_x^i & = H(c_x - \xi^i) \quad && \text{with} & \quad \xi^i & = \min(I_i),\\
    b_y^j & = H(c_y - \eta^j) \quad && \text{with} & \quad \eta^j & = \min(J_j).
  \end{aligned}
\end{equation}
That is, the activation of the BSL units depends on a threshold, implemented here as a synaptic projection from an always-active bias unit, that is defined as the minimum of the intervals $I_i$ and $J_j$ respectively for the $b_x^{i}$ and $b_y^{i}$ units. This has the effect of centering the threshold towards the left boundary of each interval (i.e. a bias of $-\xi^i$ for $b_x^i$ unit and $-\eta^j$ for $b_y^j$). Therefore, if the read-out (i.e. encoded machine configuration) of the $c_x$ and $c_y$ units in the MCL corresponded to a point on the unit square belonging to cell $D^{i,j}$, then the $b_x^i$ unit would be triggered active as well as all units $b_x^k$ with $k<i$. The same would occur for neurons $b_y^j$ and all neural units $b_y^k$ with $k<j$.\footnote{Note that the action of the BSL could be equivalently implemented by interval indicator functions represented as linear combinations of Heaviside functions.} Upon excitation, these BSL units then synaptically project to the relevant LTL units, $(t^{i,j}_x, t^{i,j}_y)$ that are naturally inactive due to a strong inhibitory bias with magnitude $h$ (the role and value of $h$ will be clarified in the subsequent Section). Specifically, each neural unit $b_x^i$ establishes synaptic excitatory connections (with weight $\frac{h}{2}$) to all LTL units within the cells $D^{k,i}$ (i.e. $(t_x^{k,i}, t_y^{k,i})$) and also project with synaptic inhibitory connections (with weight $\frac{-h}{2}$) to all LTL units within the cells $D^{k,i-1}$ (i.e. $(t_x^{k,i-1}, t_y^{k,i-1})$), where $k=1, \ldots m$; for a graphical depiction see \autoref{fig:full_BSL}. Similarly, each neural unit $b_y^j$ projects with synaptic excitatory connections (with weight $\frac{h}{2}$) to all LTL units within the cells $D^{j,k}$ (i.e. $(t_x^{j,k}, t_y^{j,k})$) and also projects with synaptic inhibitory connections (with weight $\frac{-h}{2}$) to all LTL units within the cells $D^{j-1,k}$ (i.e. $(t_x^{j-1,k}, t_y^{j-1,k})$), where $k=1, \ldots n$; see \autoref{fig:full_BSL}. The combined effect of the $b^{i}_x$ units and $b^{j}_y$ is therefore to counterbalance through their synaptic weights the natural inhibition (of bias $h$) of the LTL units in cell $D^{i,j}$. In other words each couple of LTL units $(t^{i,j}_x, t^{i,j}_y)$ receives an input $B_x^i + B_y^j$, defined as follows:
\begin{equation}
  \begin{aligned}
    B_x^i &&=&& b_x^i \frac{h}{2} + b_x^{i+1} \frac{-h}{2}\\
    B_y^j &&=&& b_y^j \frac{h}{2} + b_y^{j+1} \frac{-h}{2},
  \end{aligned}
  \label{eq:Bi+Bj_definition}
\end{equation}
where the input sum
\begin{equation}
  B_x^i + B_y^j =
  \begin{cases}
    h &\mbox{ if } (c_x, c_y) \in D_{i,j}\\
    \frac{h}{2} & \mbox{ if } c_x \in I_i, c_y \not\in J_j \quad \mbox{ or }\quad c_x \not\in I_i, c_y \in J_j\\
    0 & \mbox{ if } (c_x, c_y) \not\in D_{i,j}
  \end{cases}
  \label{eq:Bi+Bj_value}
\end{equation}
only triggers the relevant LTL unit if it reaches the value $h$. That is, if the pair $(t^{i,i}_x, t^{i,j}_y)$, is selected by the BSL units (and thus $(c_x, c_y) \in D_{i,j}$), then $B_x^i + B_y^j = h$. Otherwise $B_x^i+B_y^j$ is either equal to $\frac{h}{2}$ or $0$. An example of this mechanism is shown in~\autoref{fig:full_BSL}, where the LTL units in cell $D^{1,2}$ are activated via mediation of $b_x= \{b^{1}_x, b^{2}_x, b^{3}_x\}$ and $b_y = \{b^{1}_y, b^{2}_y\}$. Here, both $b^{3}_x$ and $b^{2}_y$ are not excited since respectively $c_x$ and $c_y$ are not activated enough to drive them towards their threshold. However, $b_x^{2}$ excites (with weights $\frac{h}{2}$) the LTL units in cell $D^{2,2}$ and $D^{1,2}$ and inhibits (with weights $\frac{-h}{2}$) the LTL units in cell $D^{2,1}$ and $D^{1,1}$. Equally, $b_y^{2}$ excites (with weights $\frac{h}{2}$) the LTL units in cell $D^{2,1}$, $D^{2,2}$ and $D^{2,3}$ and inhibits (with weights $\frac{-h}{2}$) the LTL units in cells $D^{1,1}$, $D^{1,2}$ and $D^{1,3}$. The $b_x^{1}$ and $b_y^{1}$ units excite respectively cells $\{D^{2,1}, D^{1,1}\}$ and $\{D^{1,1}, D^{1,2}, D^{1,3} \}$, but these do not inhibit any cells (due to boundary conditions).

\subsubsection{Linear Transformation Layer}
The LTL embodies the set of affine-linear transformations  of the NDA from which the network is constructed, and thus the set of symbolic operations defined by the transition table of the simulated automaton. This endows the LTL with the functional ability of generating an updated encoded machine configuration from the current one. That is, the affine-linear transformation of a NDA, $\Phi^{i,j}(x,y) =( \lambda^{i,j}_x x + a^{i,j}_x, \lambda^{i,j}_y y+ a^{i,j}_y)$ within a cell $D^{i,j}$ is simulated by the LTL unit $(t^{i,j}_x, t^{i,j}_y)$. This induces the following mapping:
\begin{eqnarray}
  (t^{i,j}_x, t^{i,j}_y) &=& \zeta_{3}^{i,j}(x,y) = \rho(\cdot, \cdot, \Phi^{i,j}(x,y), \cdot).
\end{eqnarray}
This affine-linear transformation is implemented in the form of synaptic computation, which is only triggered when the BSL units provide enough excitation enabling the two neural units $(t^{i,j}_x, t^{i,j}_y)$ to cross their threshold value and execute the operation. The read-out of this process is as follows:
\begin{equation}
  \begin{array}{lcl}
    t^{i,j}_x &=& R(\lambda^{i,j}_x c_x + a^{i,j}_x - h + B_x^i + B_y^j)\\
    t^{i,j}_y &=& R(\lambda^{i,j}_y c_y + a^{i,j}_y - h + B_x^i + B_y^j),
  \end{array}
  \label{eq:LTL}
\end{equation}
that is, initially the LTL units are rendered inactive with a strong inhibition bias $h$ implemented as a synaptic projection from a bias unit, which is defined as follows:
\begin{equation}
  - \frac{h}{2} \leq -\max_{i,j,k}(a^{i,j}_k + \lambda^{i,j}_k) \quad \text{with} \quad k=\{x,y\}.
  \label{eq:inhibition}
\end{equation}\\
This results from the fact that each BSL inputs $B^{i}_x$ and $B^{i}_y$ contribute respectively to half of the necessary excitation ($\frac{h}{2}$), that sum up and counterbalance the LTL's natural inhibition (refer to \autoref{eq:Bi+Bj_definition} and~\autoref{eq:Bi+Bj_value}). The LTL units also receive inputs from the MCL units $(c_x, c_y)$, which are respectively modulated by the synaptic weights $(\lambda^{i,j}_x, \lambda^{i,j}_y)$ and once the LTL units cross their threshold (mediated by the ramp activation function) then the intrinsic constant LTL neural dynamics $(a^{i,j}_x, a^{i,j}_y)$ completes the desired affine-linear transformation. The read-out is an updated encoded machine configuration, which is then synaptically projected back to the MCL units $(c_x, c_y)$, initiating the next computation step (related to the original machine).

\subsection{Neuronal Observation Models}
\label{sec:neural_observation_models}

In order to compare connectionist simulation results with experimental evidence from neurophysiology or psychology, one needs a mapping from the high-dimensional neural activation space $\Gamma \subset \mathbb{R}^n$ into a much lower-dimensional \emph{observation space} that is spanned by $p \in \mathbb{N}$ observables $\varphi_k: \Gamma \to \mathbb{R}$ ($1 \le k \le p$). A standard method for such a projection is PCA \citep{Elman91}. If PCA is restricted to the first principal axis, the resulting scalar variable could be conceived as a measure of the overall activity in the neural network (as in \citealp{graben2008towards}). Other important scalar observables that have been discussed in the literature are Smolensky's harmony \citep{Smolensky86} 
\[
 H = \sum_{ij} u_i w_{ij} u_j
\]
with $\vec{u} = (u_i)$ as  the network's activation vector and $\vec{W} = (w_{ij})$ its synaptic weight matrix, or Amari's mean network activity \citep{Amari74}
\begin{equation}\label{eq:amari}
    A = \frac{1}{n} \sum_{i} u_i \:.
\end{equation}
The development of biophysically inspired observation models is an important research field in computational neuroscience \citep{GrabenRodrigues13} as it could eventually lead to ``synthetic'' local field potentials (LFPs), electroencephalogram (EEG), or event-related brain potentials (ERPs) \citep{barres2013synthetic}. We shall use Amari's measure (\ref{eq:amari}) to derive such synthetic ERPs in what follows.

\section{Results}
The implementation of the R-ANN discussed in the previous Sections simulates a NDA in real-time and thus simulates its associated machine in real-time. More formally, it can be shown that under the map $\rho(\cdot)$ the commutativity property, $\zeta \circ \rho = \rho \circ \Phi$ (see commutative diagram of \autoref{fig:complete_mapping}) is satisfied. The NDA simulation (and thus the machine simulation) by the R-ANN is achieved by a combination of synaptic and neural computation among three neural types (MCL, BSL, and LTL) and with a total of neural units equal to
\begin{equation}
  n_\text{units} = 2 + n_\alpha + n_\beta + 2 n_\alpha n_\beta + 1
  \label{eq:n_units}
\end{equation}
where $n_\alpha$ and $n_\beta$ are the number of sub-sequences that can appear respectively in the left and right Domain of Dependence of the VS from which the NDA and the R-ANN are constructed. That is, a total of 2 MCL units, $(n_\alpha + n_\beta)$ BSL units, $2 n_\alpha n_\beta$ LTL units and a bias unit, that establish synaptic connections according to a synaptic weight matrix $\mathcal{A}$ of size $(n_\text{units} \times n_\text{units})$ following the connectivity pattern described in~\autoref{fig:network_architecture}. Specifically, the synaptic weights in $\mathcal{A}$ are entries from the set $\{0, 1, \frac{h}{2}, \frac{-h}{2}\} \cup \{a^{i,j}_k - h \; \vert \; i=1,\ldots,n_\alpha n_\beta \enskip,\enskip j=1,\ldots,n_\beta \enskip, \enskip k=x,y\}$, with the second set being the set of biases. A point worth mentioning is that the original formulation of the NDA relied on a simple G\"{o}del encoding of the machine configurations, but subsequent work highlighted the advantages of using a more flexible representation by employing Cylinder sets, in order to preserve important structural relationships of the symbolic descriptions and to facilitate modeling~\citep{graben2009inverse, graben2008towards, graben2004language}. Our R-ANN can be extended to incorporate a Cylinder set encoding of machine configurations by simply doubling the MCL and LTL layer.

An important modeling issue to consider is that of the halting conditions for the ANN, i.e. when to consider the computation as terminated. VSs, on which NDA and consequently our ANN model depend, do not define explicit halting conditions. However, two equally reasonable choices of halting conditions could be employed as follows. The first one is that of using a \emph{homunculus} \citep{graben2004language}, an external observer which decides to intervene on the computation once some condition is met (for example, halting the computation when the input is in a certain region of the unit square). The second one is that of using a fixed point condition: implementing a machine halting state as an Identity branch on the NDA. This way a halting configuration will result in a fixed point on the NDA, and thus on the R-ANN. In other words, the network's computation halts if and only if
\begin{equation}
  \zeta_1(x',y') = (x', y').
  \label{eq:halting_condition}
\end{equation}
A halting by \emph{homunculus} could be more appropriate in the context of interactive computation~\citep{graben2008towards, Wegner98} where constant and non-terminating interaction with the environment is assumed, or in cognitive modeling, where different kinds of fixed points, either desired or unwanted ones, are required in order to describe sequential decision problems~\citep{RabinovichEA08}, such as linguistic garden paths~\citep{graben2004language, graben2008towards}.

We will now present two examples to demonstrate the strength of our developed methodology in mapping automata computation to R-ANN computation in real-time (an additional example on Turing Machines is available in the supplementary materials). The source code for all the examples is freely accessible via~\citet{TMtoNDsource}. 

\subsection{Example 1: Finite-State Locomotive Pattern Generator}
FSMs are at the basis of many state-of-the-art approaches to the construction of locomotion controllers for articulated robots (see for example \citealp{alvarez2012human, collins2005bipedal}). They are easy to design, implement, and debug, and their relation with animal gait is well characterized \citep{mcghee1968some}. On the other hand, recent research in robot locomotion control shows an increasing interest towards alternative approaches based on CPGs, neural networks capable of producing rhythmic patterns of activation in absence of rhythmic input sources. In his 2008 paper, \citeauthor{ijspeert2008central} presented the benefits and drawbacks of CPGs with respect to other approaches for robot locomotion control. We briefly summarize the benefits identified by the author: i) the rhythmic behavior supported by CPGs is robust to the transient perturbation of state variables; ii) CPGs are well-suited for distributed implementations (such as in modular robots); iii) CPGs reduce the dimensionality of the control problem by introducing few high-level control parameters allowing for the modulation of the locomotion; iv) CPGs are ideally suited for the integration of sensory feedback through coupling terms in the differential equations of the controller; v) CPGs often work well with learning and optimization algorithms.
On the other hand, as specified by the author, CPG-based approaches are still lacking of a sound design methodology and theoretical grounding for their description. In the example presented in this Section, we will show how our mapping could aid the design of CPGs producing arbitrary patterns for locomotion in robots, starting from a FSM description of the desired rhythmic pattern. By combining the two approaches, the design of these controllers benefits from the solid theoretical grounding of FSM-based locomotion and from its ease of design and implementation. To contextualize our derived CPG in terms of familiar animal locomotion, we qualitatively model the results of a well-known experiment on cat gait.

In their seminal work, \citet{shik1966control} applied different levels of electrical stimulation to the midbrain of a decerebrated cat. The authors observed transitions in the gait of the animal as an increasing level of stimulation was applied, eliciting first a {\it walk}, then a {\it trot} and finally a {\it gallop} gait. Our theoretical framework can qualitatively reproduce these experimental observations, by deriving a R-ANN which generates the relevant gait patterns, and reproduces the transition between them as a function of the applied stimulus strength. To keep the exposition simple, we will only consider the {\it walk} and {\it gallop} gaits, and the transition between the two. In the study of the mammalian quadruped gait, the four legs are numbered so that each gait can be associated with a certain sequence, given by the order in which the legs touch the ground over one gait cycle. The left and right hind legs are associated respectively with the numbers $1$ and $2$, and the left and right fore legs are associated respectively with the numbers $3$ and $4$. The gait cycle is assumed to start when the left hind leg touches the ground. A {\it walk} gait is thus defined by the sequence $(1, 3, 2, 4)$, and a {\it gallop} gait is defined by the sequence $(1, 2, 3, 4)$. At a very high level, the computation carried out by the CPG in charge of producing the gait patterns in the quadruped mammalian can be informally stated as: if stimulation from midbrain is low, sequentially activate legs following pattern $(1, 3, 2, 4)$. If it is high, sequentially activate legs from pattern $(1, 2, 3, 4)$. We can implement the low level and high level of stimulation as the two input symbols of a FSM, and construct the $\delta$ transition function to sequentially reproduce the two patterns by switching between states. The FSM can thus be defined as in~\autoref{tab:delta_table_CPG}.\\
\begin{table}\centering
  \ra{1.3}
  \begin{tabular}{@{}rccccc@{}}\toprule
    Symbols &  & \multicolumn{4}{c}{States} \\
    \cmidrule{3-6}
    & & $q_1$ & $q_2$ & $q_3$ & $q_4$\\ \midrule
    \texttt{<lo>} & & $q_3$ & $q_4$ & $q_2$ & $q_1$ \\
    \texttt{<hi>} & & $q_2$ & $q_3$ & $q_4$ & $q_1$ \\
    \bottomrule
  \end{tabular}
  \caption{{\bf State transition table for the simulated Central Pattern Generator finite-state automaton.} It is possible to observe how different input leads to different produced patterns, implemented as sequences of states.}
  \label{tab:delta_table_CPG}
\end{table}
This FSM can now be mapped (via our proposed approach) into a R-ANN, consisting in this case of $22$ neural units (according to~{\autoref{eq:n_units}). The chosen gamma functions for the G\"{o}del encoding of this FSM are defined as follows:
\begin{align*}
  \gamma_s(\sigma) & := \begin{cases}
    0 &\mbox{ if } \sigma = \texttt{<lo>} \\
    1 &\mbox{ if } \sigma = \texttt{<hi>} \\
  \end{cases} &
                \gamma_q(q) & :=
                              \begin{cases}
                                0 &\mbox{ if } q = q_1 \\
                                1 &\mbox{ if } q = q_2 \\
                                2 &\mbox{ if } q = q_3 \\
                                3 &\mbox{ if } q = q_4 \\
                              \end{cases}
\end{align*}
The step-by-step dynamics of the derived R-ANN can be observed in~\autoref{fig:cpg_sim}. Here we use the machine's input as the substrate for the external stimulus, which is ultimately encoded by the neural unit $c_y$ within our R-ANN as shown in the bottom plot of \autoref{fig:cpg_sim}. Note how we manipulate the activation of $c_y$ to gradually increase from a low to a high level of stimulation. That is, we introduce a continuous control parameter into an originally pure symbolic model, enabling us to carry out a bifurcation study in analogy with traditional coupled oscillator models \citep{GolubitskyStewartEA99, GolubitskyStewartEA98, SchonerJiangKelso90, CollinsRichmond94}. Under this stimulation, the R-ANN defined by the mapping qualitatively reproduces the key features of the CPG involved in the locomotion and transitions described in \citet{shik1966control}. In particular, it is possible to observe how low levels of stimulation elicit the production of the {\it walk} gait cycle, whereas an increase in the level of stimulation induces a sudden transition to  the {\it gallop} gait cycle.

This key relation between the stimulation level (i.e a real control parameter) and the computation carried out by the network, which can be related to the underlying symbolic space thanks to the mapping, depends upon an informed decision in the gamma numbering of the states for the G\"{o}del encoding.
In fact, the chosen gamma numbering ensures that the unit square encoding of machine configurations where \verb|<lo>| is the current input symbol corresponds to all points $(x, y)$ such that $x < \psi_y(\texttt{<hi>)}$ where $\psi_y$ is defined as in \autoref{eq:godel_encoding}, and specifically $\psi_y(\texttt{<hi>}) = \gamma_s(\texttt{<hi>}) g_s^{-1}= \frac{1}{2}$. In terms of the underlying NDA representation, increasing the activation of $c_y$ until its value reaches and exceeds $\frac{1}{2}$ corresponds to forcing the encoded machine state to cross the boundary between cells associated to a \verb|<lo>| input symbol to those associated to a \verb|<hi>| input symbol, thus causing a transition between a \emph{walk} and a \emph{gallop} gait. Note that in this example, we do not model halting conditions for the derived network, as it is not clear what halting means in the context of the computation performed by CPGs.

\begin{figure}
\includegraphics[width=\linewidth]{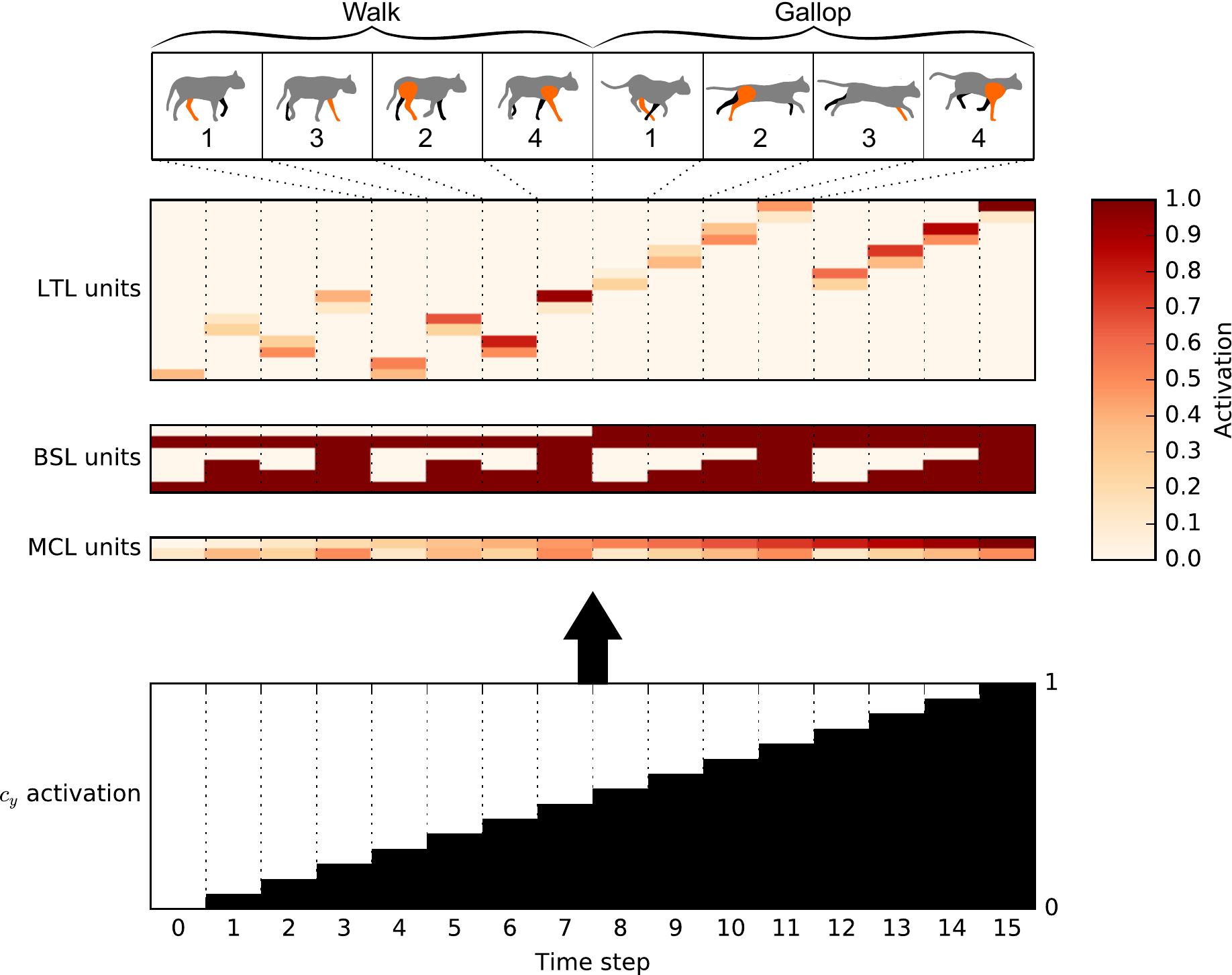}
\caption{{\bf Recurrent artificial neural network functioning as a Central Pattern Generator.} The network reproduces the qualitative behavior of the locomotive Central Pattern Generator described in \citet{shik1966control}. In the bottom plot, the level of stimulation applied to the network through neuron $c_y$ is shown. In the top three plots, the levels of activation of each neural unit in the three layers is shown for each time step. Note how two different patterns, {\it walk} and {\it gallop}, are generated depending on the level of stimulation. This results from the way the original finite-state machine was programmed.}
  \label{fig:cpg_sim}
\end{figure}
To summarize, we derived a CPG from a FSM description of a locomotion controller, inspired by results on the generation of gait patterns in the cat midbrain. By doing so, we outlined a new design methodology for CPG-based locomotion control in robots which does not suffer from some of the drawbacks of other CPG approaches, by grounding the description and design of the CPG on the theoretical grounding of FSM-based approaches. Some problematic aspects of the methodology we outlined are due to the discrete-time nature of our mapping. In fact, fully realizing the benefits of CPG-based approaches summarized at the beginning of this Section requires continuous time models. This notwithstanding, we believe that the proof of concept we provide here already shows encouraging results for future developments.

As an additional remark, the methods we describe in this paper are ideally suited for the deriving of neural networks implementing paradigms of interactive computation, as we will demonstrate shortly. This is especially relevant for the design of CPGs. In fact, recent research has unveiled a surprising degree of hierarchical organization in mammalian respiratory CPGs, which allows for a highly robust and flexible pattern production that can adapt to a variety of conditions (see for example work by \citealp{smith2013brainstem, smith2007spatial}). Our methodology easily accommodates the mapping of hierarchies of automata to hierarchically organized neural networks, as we demonstrate in the next example through the modeling of garden-path parsing, a concept employed in language processing \citep{graben2004language}. Importantly, networks of automata could be used to design complex pattern generation in modular robots (see \citealp{sprowitz2014roombots} for a recent example of modular robots using a distributed CPG for locomotion).

\subsection{Example 2: Interactive Automata Networks}
\label{sec:garden_path}

Interactive computation \citep{Wegner98} is a recent theoretical development that seeks to formalize the complexity of interactions that we observe in real-world computing. In classical Automata Theory, the interaction between an automaton and the external world is restricted to an input-output relation. That is, the external world provides an input, the automaton performs its computation on that input, and then returns an output to the external world. Within the framework of interactive computation, instead, automata can interact with the external world (and with other automata) at every step of their computation. External forces can act on the configuration of the automaton, and the configuration can itself affect the external world. Clearly, this framework provides a much richer language to describe models of computation, and is especially useful to express notions of compositionality and concurrency. These constructs are essential not only in the study of modern computing systems, but also in the context of cognitive modeling.
In this example, we will build a model of the human processing of locally ambiguous sentences by constructing a network of interactive automata. Through this proof-of-concept, we want to demonstrate the flexibility of our approach by showing how it can be seamlessly used to construct neural networks implementing interactive systems. In order to do so, we choose a system that i) is simple enough to allow for clear exposition, but complex enough to carry out a meaningful computation; ii) is composed by a range of different automata; iii) incorporates different forms of interaction between its automata  components.

Garden-path sentences are locally ambiguous sentences that induce the temporary production of an erroneous parse by the reader, which is then forced to reconsider their interpretation of the previously presented material in order to finally reach a correct parse. Consider for example the sentence ``I convinced her children are noisy''. In reading the sentence, the reader first constructs an intermediate parse where ``her children'' is the object of the phrase ``I convinced''. After reading the rest of the sentence, the reader realizes that the intermediate parse was incorrect: ``her'' is the object of ``I convinced'', and ``children are noisy'' is a subordinate clause. The reader thus reanalyzes the sentence to produce a correct parse. \citet{OsterhoutHolcombSwinney94} have shown that the reanalysis of a sentence due to a garden-path is associated in the brain of the reader with a positive deflection 600 milliseconds (P600) after the onset of a garden-path -- the word ``are'' in the example above -- in sequentially presented sentences, as measured by a trial averaged electroencephalogram (thus obtaining event-related brain potentials).

Many proposals have been advanced to account for the mechanisms underlying the reanalysis of incorrectly parsed sentences due to garden-path effects. In our model, we implement the reanalysis through a diagnosis and repair mechanism, described in \citet{lewis1998reanalysis}. By this account, the parser tries to incrementally build a parse as the sentence material is presented. If a dead-end is reached (i.e. the parser becomes stuck in a garden-path), the parser diagnoses the need for reanalysis, and the search space of possible continuations of the parse is modified by some repair operator that ``bridges'' the dead-end to another point in the search space, allowing the parser to correctly complete the processing of the sentence.
The parser model we create implements this mechanism  to process garden-path sentences where the local ambiguity is given by the incorrect assignment of the subject and object grammatical constituents.

In many languages, native speakers have been shown to prefer to interpret an ambiguous nominal constituent as a subject rather than an object. 
Consider for example the following two sentences, extracted from the ERP study on ambiguous pronouns by \citet{FrischGrabenSchlesewsky04} on German speakers. 
Both sentences start with
\begin{center}
  \fbox{
    \begin{minipage}{\linewidth}
      { \small
        \begin{tabbing}
          \; \= \textit{Nachdem} \; \= \textit{die Kommissarin} \; \= \textit{den Detektiv} \; \= \textit{getroffen hatte} \; \= \textit{\ldots}\\
          \; \= After \> the cop \> the detective \> had met \> \ldots \\
          \rule{.5\linewidth}{.1em}\\
          ``\>After the cop had met the detective, \ldots''
        \end{tabbing}
      }
    \end{minipage}
  }
\end{center}
One of the sentences then continues with a clause in subject-object order (s-o sentence), i.e. the preferred order in the parsing of ambiguous constituents: %
\begin{enumerate}[(1)]
  \small
  \itemsep2em
\item[(s-o sentence)]
  \begin{center}
  \fbox{
    \begin{minipage}{\linewidth}
      \begin{tabbing}
        \ldots \; \= \textit{sah} \; \= $\text{\textit{sie}}_\textbf{s}$ \; \= $\text{\textit{den Schmuggler}}_\textbf{o}$ \\
        \ldots \; \> saw \> she \> the smuggler\\
        \rule{.5\linewidth}{.1em}\\
        \ldots \; \> ``she saw the smuggler''
      \end{tabbing}
    \end{minipage}
  }
\end{center}
\end{enumerate}
In this case, the reader correctly interprets ``sie'' to be the subject of the second clause, and ``den Schmuggler'' as the object (as ``den Schmuggler'' is in the accusative case, thus specifying a direct object to the verb ``sah'').\\
The second sentence is instead in the dispreferred object-subject order (o-s sentence):
\begin{enumerate}[(1)]
  \small
  \itemsep2em
\item[(o-s sentence)]
  \begin{center}
  \fbox{
    \begin{minipage}{\linewidth}
      \begin{tabbing}
        \ldots \; \= \textit{sah} \; \= $\text{\textit{sie}}_\textbf{o}$ \; \= $\text{\textit{der Schmuggler}}_\textbf{s}$ \\
        \ldots \; \> saw \> she \> the smuggler\\
        \rule{.5\linewidth}{.1em}\\
        \ldots \; \> ``the smuggler saw her''
      \end{tabbing}
    \end{minipage}
  }
  \end{center}
\end{enumerate}
The psycholinguistic study by \citet{FrischGrabenSchlesewsky04} has shown that the reader first tries to apply the preferred subject-object parsing strategy to this clause (and sentences with similar subject/object pronoun ambiguity). The reader thus initially interprets ``sie'' as the subject of the clause in nominative case, expecting it to be followed by the object in accusative. Upon further reading, however, they realize that ``der Schmuggler'' is in the nominative case instead, and thus has to be the subject. This leads the reader to reconsider the previous material to correctly parse ``sie'' as a pronoun in accusative case, the direct object of the verb ``sah''. This reanalysis was observed as a P600 effect in the ERP.
\\
At a high level of abstraction \citep{graben2004language}, we can capture the structure of these sentences through a CFG $G$ with production rules:
\begin{align}
  &\texttt{S} \rightarrow \texttt{s o} \tag{\textit{s-o}} \label{eq:rule_so} \\
  &\texttt{S} \rightarrow \texttt{o s}, \tag{\textit{o-s}} \label{eq:rule_os}
\end{align}
where \texttt{S} is a distinguished starting non-terminal, and where the \texttt{s} and \texttt{o} terminals stand respectively for ``subject'' and ``object'' phrase.

In our model we thus split the $G$ grammar into two grammars $G_\textit{s-o}$ and $G_\textit{o-s}$, comprising respectively of the \ref{eq:rule_so} and \ref{eq:rule_os} production rules \citep{graben2004language}, and reflecting the existence of two strategies in the parsing of sentences with subject/object pronoun ambiguity. To recognize the two different sentence structures, our model is endowed with two specialized TDRs, constructed from the $G_\textit{s-o}$ and $G_\textit{o-s}$ grammars as shown in \autoref{sec:TDR}.
Initially, the \textit{s-o} TDR is tried on the input, to model the subject-object interpretation preference. In case it fails because of a garden path, the model acts as prescribed by a diagnosis and repair account. That is, it first diagnoses that a problem has arisen in parsing, repairs the parse, and finally switches strategy to correctly parse the input.
In order to implement the diagnosis step, our model needs a way to monitor the state of the parse and extract the relevant diagnostic information. We implement this through a \textit{Diagnosis} PDA (see \autoref{tab:diagnosis_PDA}), which compares the current parse with that from the previous time step; if the parse didn't change, that means that the parser is stuck and can't process the input further. In that case the \textit{Diagnosis} PDA changes its state to an ``error'' state, thus implementing the diagnosis step.
\begin{table}\centering
  \ra{1.3}
  \begin{tabular}{@{}rcr@{\;}c@{\;}lr@{\;}c@{\;}lr@{\;}c@{\;}l@{}}\toprule
    Symbols &  & \multicolumn{9}{c}{States} \\
    \cmidrule{3-11}
    & & \multicolumn{3}{c}{$\pss{pda}{q}{idle}$} & \multicolumn{3}{c}{$\pss{pda}{q}{parsing}$} & \multicolumn{3}{c}{$\pss{pda}{q}{error}$}\\ \midrule
    $(\sqcup, \sqcup)$ & & $(\pss{pda}{q}{idle}$&,&$\sqcup)$ & $(\pss{pda}{q}{idle}$&,&$ \sqcup)$ & $(\pss{pda}{q}{idle}$&,&$ \sqcup)$\\
    $(x, y)$ & & $(\pss{pda}{q}{parsing}$&,&$y)$ & $(\pss{pda}{q}{parsing}$&,&$ y)$ & $(\pss{pda}{q}{parsing}$&,&$ y)$\\
    $(x, x); \> x\ne\sqcup$ & & $(\pss{pda}{q}{error}$&,&$x)$ & $(\pss{pda}{q}{error}$&,&$ x)$ & $(\pss{pda}{q}{error}$&,&$ x)$\\
    \bottomrule
  \end{tabular}
  \caption{{\bf State transition table for the \textit{Diagnosis} push-down automaton (PDA).} The input to this machine is the parse produced by the top-down recognizers (TDRs). For any state, input symbol and stack symbol, the machine pushes its input to the stack, in order to be able to compare its current input with the one from the previous time step. In particular, if the input symbol and top-of-stack are blank symbols, the machine transitions to an ``idle'' state, signaling that nothing is happening; if the current input and the one from the previous time step are different, the machine transitions to a ``parsing'' state, signaling that the TDRs are successfully parsing their input; if the current input and the one from the previous time step are the same (but not both blanks), then the TDR parsing the input is stuck, and the machine transitions to an ``error'' state.}
  \label{tab:diagnosis_PDA}
\end{table}
The repair step is realized by introducing a \textit{Repair} VS , that can be described by the following rewriting rule:
\begin{equation}
  \texttt{s o}\; . \; w \rightarrow \texttt{o s}\; . \; w,
\end{equation}
corresponding to a reanalysis of the ambiguous sentence in terms of the dispreferred object-subject sentence structure. Once the sentence has been reanalyzed and thus the parse repaired, the second parser can proceed to process the input until it has been completely consumed and the stack is emptied.
In order to switch strategies, our model needs a higher-level controller that has access to diagnostic information about the current parse, and decides which parsing strategy to apply. In particular, this controller should first activate the preferred \textit{s-o} TDR. If the parser failed (as signaled by the \textit{Diagnosis} PDA) then the higher-level controller should first activate the \textit{Repair} VS to allow for the reanalysis of the ambiguous sentence, and subsequently activate the \textit{o-s} TDR. We implement the high level controller through a \textit{Strategy} FSM (see \autoref{tab:strategy_FSM}), endowed with the capability of selectively activating the \textit{s-o} and \textit{o-s} TDRs, as well as the \textit{Repair} VS, by switching its internal state. This machine receives the diagnostic information provided by the \textit{Diagnosis} PDA as input.
\begin{table}\centering
  \ra{1.3}
  \begin{tabular}{@{}rcccc@{}}\toprule
    Symbols &  & \multicolumn{3}{c}{States} \\
    \cmidrule{3-5}
    & & $\pss{fsm}{q}{s-o}$ & $\pss{fsm}{q}{o-s}$ & $\pss{fsm}{q}{repair}$\\ \midrule
    $\pss{pda}{q}{idle}$ & & $\pss{fsm}{q}{s-o}$ & $\pss{fsm}{q}{s-o}$ & $\pss{fsm}{q}{s-o}$ \\
    $\pss{pda}{q}{parsing}$ & & $\pss{fsm}{q}{s-o}$ & $\pss{fsm}{q}{o-s}$ & $\pss{fsm}{q}{o-s}$ \\
    $\pss{pda}{q}{error}$ & & $\pss{fsm}{q}{repair}$ & $\pss{fsm}{q}{o-s}$ & $\pss{fsm}{q}{o-s}$ \\
    \bottomrule
  \end{tabular}
  \caption{{\bf State transition table for the \textit{Strategy} finite-state machine (FSM).} The input to this machine is the diagnostic information produced by the \textit{Diagnosis} push-down automaton (PDA), i.e. its state. The FSM starts in state $\pss{fsm}{q}{s-o}$. In fact, the preferred parsing strategy is that implemented by the \textit{s-o} top-down recognizer (TDR), corresponding to the parsing of subject-object sentences, so that it is tried first. If the \textit{s-o} TDR fails, the \textit{Diagnosis} PDA signals an error; the input sentence is not in subject-object order, and a switch of parsing strategy is needed. The \textit{Strategy} FSM first changes state to $\pss{fsm}{q}{repair}$, activating the \textit{Repair} versatile shift (VS) so that the switch can take place. Repairing the parse leads the \textit{Diagnosis} PDA to signal that the parsing started again, so that the new input for the \textit{Strategy} FSM becomes again $\pss{pda}{q}{parsing}$. Given $\pss{pda}{q}{parsing}$ in input and $\pss{fsm}{q}{repair}$ as a current state, the FSM moves to the $\pss{fsm}{q}{o-s}$ state, leading to the activation of the \textit{o-s} TDR, until the input has been parsed.}
  \label{tab:strategy_FSM}
\end{table}
The FSM has three states, namely an ``s-o'' state, a ``repair'' state, and an ``o-s'' state. By switching between these states, the FSM can activate the respective automata.
Note that this form of interaction is not defined for the VS introduced in \autoref{sec:VS}. That is, we do not define a way for a VS to ``call'' other shifts. Extending VSs to incorporate notions of compositionality and concurrency will allow the refining of the mapping presented in this paper to reflect these new capabilities. For the moment, we just want to demonstrate the possibilities opened by the present work; for this reason, we will implement the ``subroutine'' capability in our neural network through a familiar mechanism already encountered in the previous Sections, ignoring momentarily the missing theoretical details and leaving their definition for future work.

To avoid race conditions, at most one automaton in the interactive network can re-write symbols in a sub-sequence at any given computation step . The ``parse'' sub-sequence can only be read, but not re-written, by the \textit{Diagnosis} PDA. Similarly, the ``diagnosis'' sub-sequence can only be read, but not re-written, by the \textit{Strategy} FSM. Furthermore, the selective activation of the \textit{s-o} TDR, the \textit{o-s} TDR, and the \textit{Repair} VS operated by the \textit{Strategy} FSM ensures that at any given computation step only one between these automata can perform symbolic re-writing on the ``input'' and ``parse'' sub-sequences.
\begin{figure}
  \centering
\includegraphics[width=.7\linewidth]{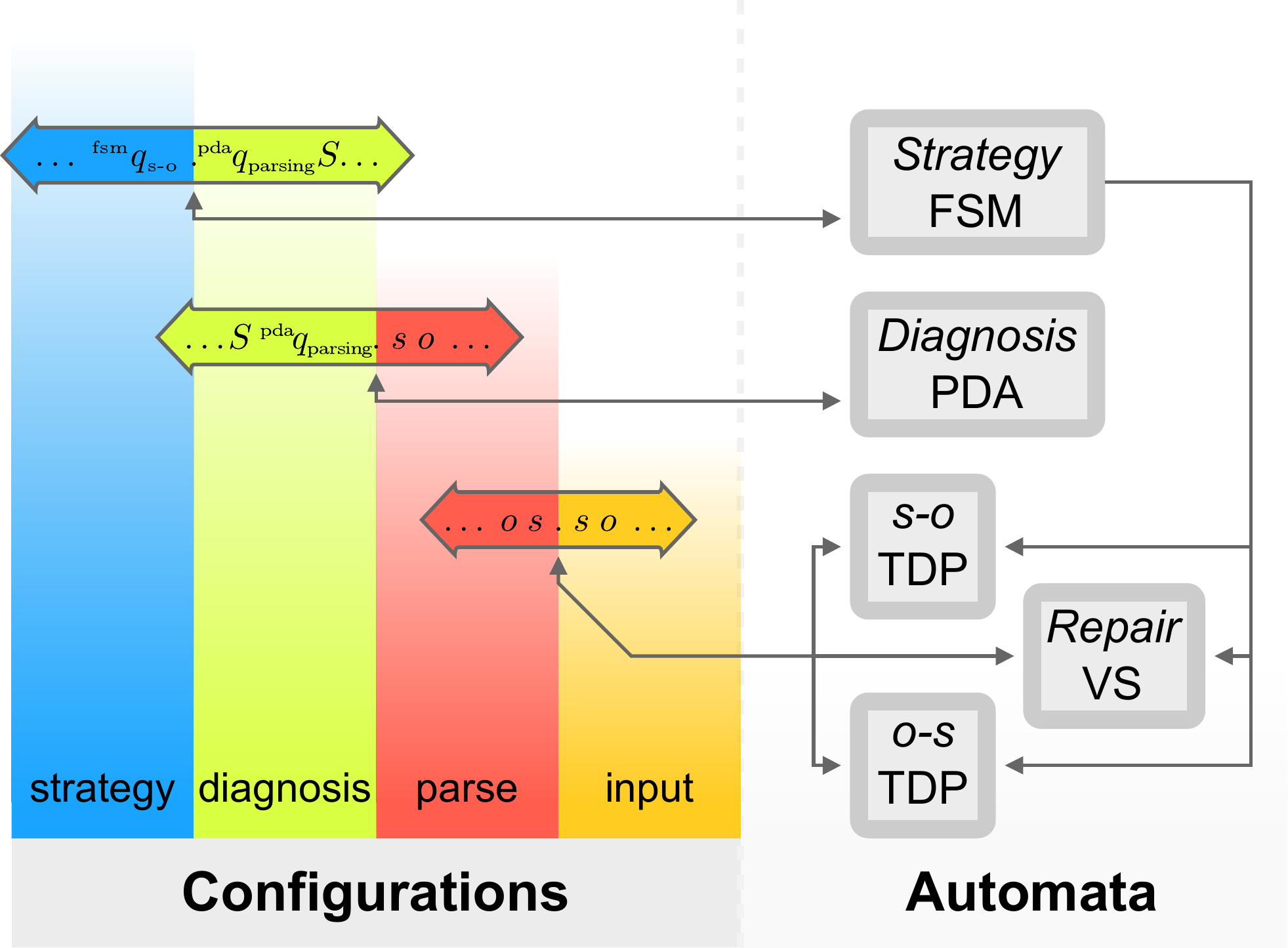}
\caption{{\bf Interactive automata network for parsing of garden path sentences.} The figure shows the complete system described in \autoref{sec:garden_path}. For simplicity, we show the various automata components as acting on their configurations represented as dotted sequences. Dotted sub-sequences of the same color are coupled, i.e. they are for all intents and purposes the same sub-sequence. For example, the ``parse'' dotted sub-sequence that contains the current stack of the top-down recognizers (TDRs) and of the \textit{Repair} versatile shift (VS), is at the same time the input tape of the \textit{Diagnosis} push-down automaton (PDA). Similarly, the ``diagnosis'' sub-sequence that stores the current state and stack of the \textit{Diagnosis} PDA, is at the same time the input tape of the \textit{Strategy} finite-state machine (FSM). Note that a second form of interaction, other than that allowed through the sharing of dotted sequences, is present in the automaton. In particular, the \textit{s-o} and \textit{o-s} TDRs and the \textit{Repair} VS are activated based on the state of the \textit{Strategy} FSM. }
  \label{fig:interactive_automaton}
\end{figure}

To map the system of interactive automata to a R-ANN, we first convert each of its component in the familiar way, as described in the previous Sections. That is, the \textit{s-o} and \textit{o-s} TDRs, the \textit{Repair} VS, the \textit{Diagnosis} PDA, and the \textit{Strategy} FSM are first converted to VSs acting on dotted sequences, then mapped to their NDA representation and finally to R-ANNs.
The G\"{o}delizations of the ``input'', ``parse'' and ``strategy'' sub-sequences are defined as in \autoref{eq:godel_encoding}, with each of the gamma enumerating functions defined as follows:
\begin{equation}
\begin{aligned}
  \gamma_\text{input} & := \{(\sqcup, 0), (\texttt{S}, 1), (\texttt{o}, 2), (\texttt{s}, 3)\}\\
  \gamma_\text{parse} & := \{(\sqcup, 0), (\texttt{o}, 1), (\texttt{s}, 2)\}\\
  \gamma_\text{diagnosis} & := \{(\pss{pda}{q}{idle}, 0), (\pss{pda}{q}{parsing}, 1), (\pss{pda}{q}{error}, 2)\}
                           \label{eq:gammas_interactive}
\end{aligned}
\end{equation}
where each function is represented as a set of $(\sigma, k)$ pairs, with $\sigma$ being a symbol and $k\in\mathbb{N}$ its enumeration.
The G\"{o}delization of the ``diagnosis'' sub-sequence is instead defined as in \autoref{eq:refined_encoding}, with
\begin{equation*}
  \gamma_\text{strategy} := \{(\pss{fsm}{q}{s-o}, 0), (\pss{fsm}{q}{o-s}, 1), (\pss{fsm}{q}{repair}, 2)\}
\end{equation*}
enumerating the states of the \textit{Diagnosis} PDA, and $\gamma_\text{parse}$ (already defined in \autoref{eq:gammas_interactive}) enumerating its stack symbols.
Having mapped each of the machines to a R-ANN, we can use the derived networks as components of the overall system architecture (see \autoref{fig:garden_path_net} for the full architecture). In order to simplify the exposition, we construct the overall network to feature only one set of recurrent connections. To do so, we endow our architecture with 4 Configuration Layers (CLs), containing the ``strategy'', ``diagnosis'', ``parse'', and ``input'' sub-sequences. Between each CL and the next, the network components derived from the automata are connected to perform their part of the processing on the relevant subsequences. In particular, if the VS representation of an automaton acts on some $\alpha.\beta$ dotted sequence, the input of its associated network component is connected to the units encoding the $\alpha$ and $\beta$ subsequences in the $i\text{-th}$ CL, whereas its output (which is a recurrent connection to the MCL layer in the original mapping) is connected to the units encoding $\alpha$ and $\beta$ in the $(i+1)\text{-th}$ CL.
The final CL is connected with a $\mathcal{I}_{4\times4}$ synaptic weight matrix to the first CL layer (i.e. each unit encoding a subsequence of the last CL is connected with a weight of 1 to the same unit in the first CL).
Finally, to implement the subroutine call capabilities of the strategy FSM, we add a \textit{Meta} branch selection layer that takes the ``strategy'' subsequence as input, and is connected with the lateral inhibition connection pattern specified in \autoref{sec:BSL} to the \textit{s-o} and \textit{o-s} TDRs, and to the \textit{Repair} VS. Note how this creates a nested structure, with the \textit{s-o} TDR, the \textit{o-s} TDR, and the \textit{Repair} VS functioning as higher-level symbolic operations of a \textit{Parser} machine. This is reflected in the nested structure of the \textit{Parser} R-ANN sub-network, where the lower-level machines function as cells in a LTL, controlled by the \textit{Meta} BSL (see \autoref{fig:garden_path_net}).
\begin{figure}
  \centering
\includegraphics[width=.8\linewidth]{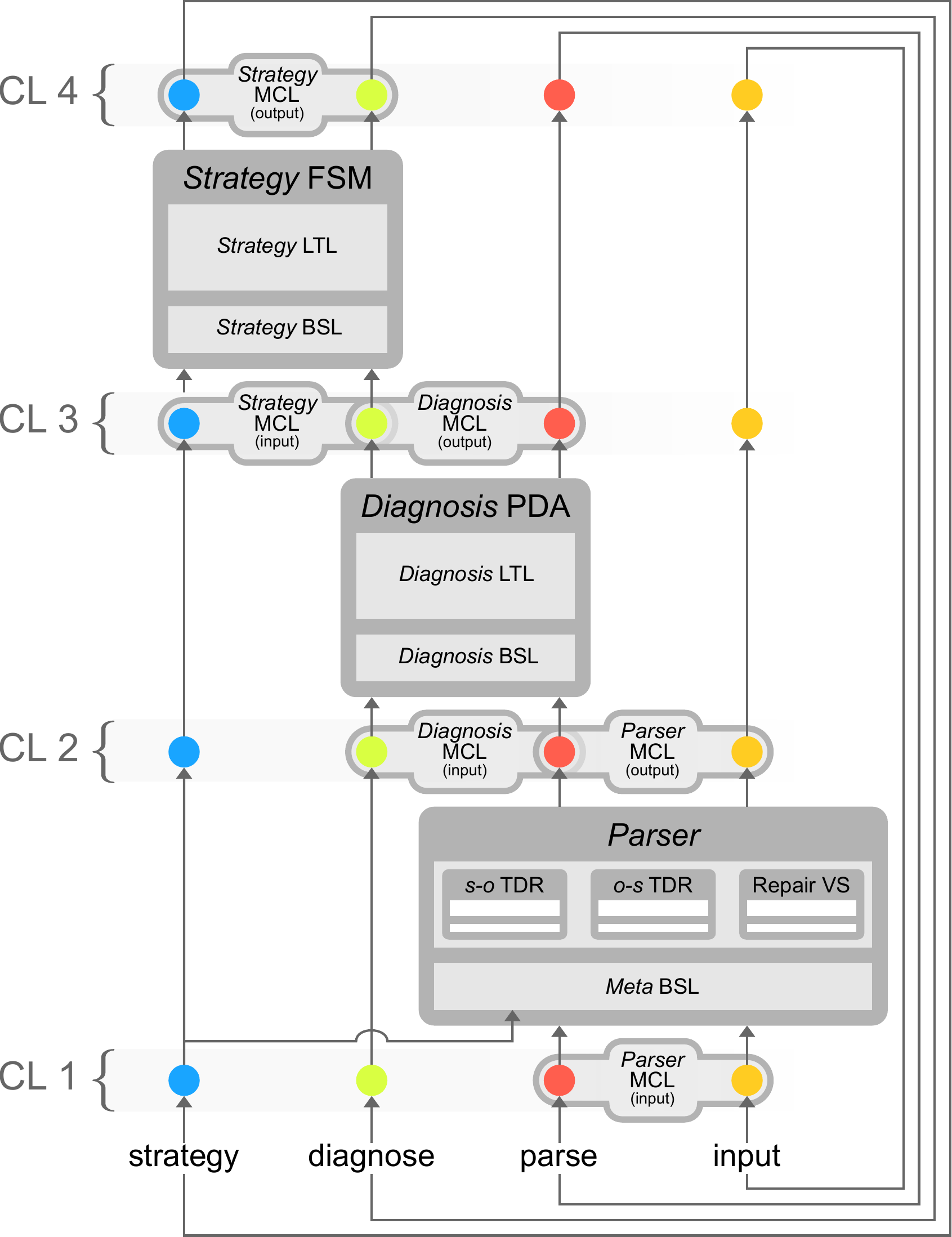}
\caption{{\bf Garden-path parsing network architecture.} In order to simplify the exposition, we construct our network such that the only recurrent connection is that from the last to the first layer of the network (i.e. CL 4 to CL 1, where CL stands for Configuration Layer). Note that the \textit{Parser} sub-network is itself composed of the \textit{s-o} top-down recognizer (TDR), the \textit{o-s} TDR, and the \textit{Repair} versatile shift (VS) sub-networks. These are arranged as cells of a linear transformation layer (LTL), and selectively activated by a \textit{Meta} branch selection layer (BSL) controlled by a ``strategy'' neural unit.}
  \label{fig:garden_path_net}
\end{figure}

In \autoref{fig:garden_path_net_acts} we show the network activation when two different sentence structures are presented in input. In particular, note the serial activation of the \textit{s-o} TDR, \textit{Repair} VS and \textit{o-s} TDR sub-networks when a object-subject sentence is presented.
\begin{figure}
  \centering
\includegraphics[width=.9\linewidth]{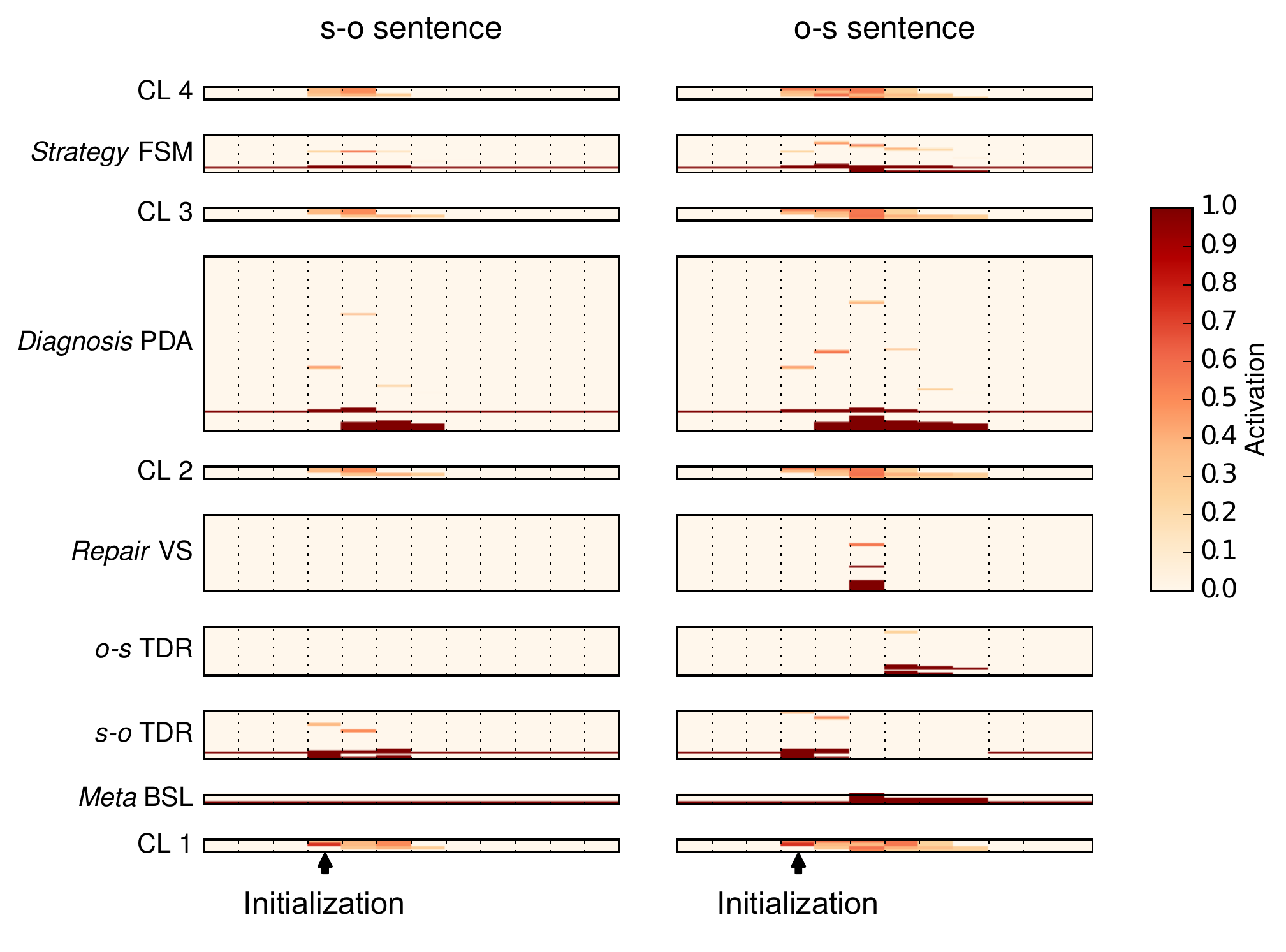}
\caption{{\bf Network activation for subject-object and object-subject sentence presentation.} Notice the serial activation of the \textit{s-o}, \textit{Repair} and \textit{o-s} sub-networks in case of an object-subject sentence presentation, and the longer ``tail'' of activation, reflecting the additional computation needed to process the dispreferred input.}
  \label{fig:garden_path_net_acts}
\end{figure}
By mapping the parser from a machine evolving in a symbolic space to a neural network evolving in a vectorial space, we are now able to compute synthetic event-related potentials, or ``synth-ERPs'', \citep{graben2008towards, barres2013synthetic} as trial-averages of the mean network activation, as discussed in~\autoref{fig:garden_path_cloud_erps}. This is achieved by calculating the mean global network activation according to \citet{Amari74} (\autoref{eq:amari}) for a simulation over 100 trials for each input stimulus, where random initial conditions compatible with the symbologram representation of the input are prepared according to~\citet{graben2008towards}. In brief, symbologram-compatible random initial conditions are generated through the G\"odelization of sequences of the form $w_\alpha u.v w_\beta$, where $u.v$ is the dotted sequence describing the input to the system, and $w_\alpha, w_\beta \in \mathbf{A}^*$ are random sequences of symbols in $\mathbf{A}$.

As~\autoref{fig:garden_path_cloud_erps} reveals, the network shows a P600-like effect in the processing of garden-path sentences, with a peak of increased and sustained activation with respect to the control condition. The simplified model of garden-path processing we presented here does not yet allow for a direct quantitative comparison with experiments such as in \citet{FrischGrabenSchlesewsky04} (in fact, a carefully crafted model would require a level of detail and attention which goes beyond the scope of this paper).
Yet, these simulations could be the starting point for more detailed statistical correlation analyses \citep{GrabenDrenhaus12, FrankEA15} in future work, relating these computations to electrophysiological measurements.
\begin{figure}
  \centering
\includegraphics[width=.8\linewidth]{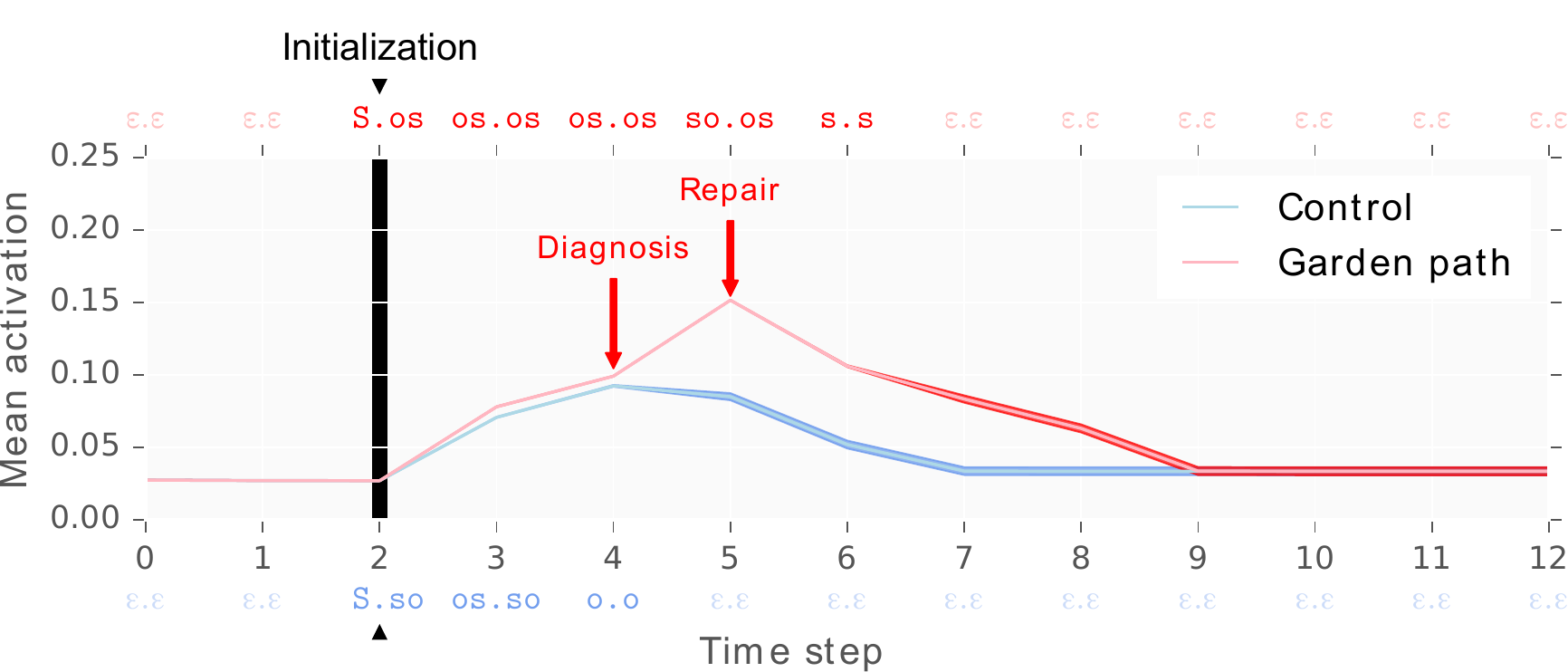}
\caption{{\bf Synthetic P600 event-related brain potential as mean Network activation for random cloud of initial conditions.} In this figure we show the mean global network activation calculated through \autoref{eq:amari} for each time step of two simulations, averaged over $100$ trials. For each of the two simulations, we run the network presenting at time $t=2$ one of $100$ random inputs generated compatibly to the symbologram representation of one of two sequences. In other words, noise is added to each input such that, if the input was generated by G\"{o}delizing a sequence of length $n$, decoding the input would yield the original sequence in the first $n$ symbols, with the rest being a random symbolic continuation. If stronger noise was added instead, that would have prevented the network to correctly perform its computation, as we would have destroyed essential input information. In blue, we show the averaged mean activation (light blue) and its standard deviation (dark blue) for a presented input encoding the sequence $\texttt{S.so}$, representing an input sequence in subject-object order, i.e. the network's preferred order as explained in \autoref{sec:garden_path}. Note that the parsing is completed at $t=5$. In red, the averaged mean activation (light red) and its standard deviation (dark red) for an input encoding the sequence  $\texttt{S.os}$, representing an input sequence in object-subject order, leading to a garden path in the parsing of the input. The time at which the diagnosis ($t=4$) and repair ($t=5$) steps are carried out in the symbolic interactive system (and thus in its recurrent artificial neural network mapping) is indicated by arrows. We also report, at the top and bottom of the plot, the configuration of the parser networks as a dotted sequence for each time step, for respectively the garden path and the control condition. Note how the garden-path processing is associated with a strong divergence in activation starting from time $t=5$, and followed by a longer tail than that of the network in the control (preferred) condition. This reflects the additional computation needed by the network to successfully resolve the garden path in parsing, and qualitatively corresponds to the P600 event-related brain potential measured in psycholinguistics experiments (see \autoref{sec:garden_path}). Furthermore, note that in both conditions the network starts and returns to a ``resting state'', waiting for input to process from the external world, implementing a notion of continuous computation which is the hallmark of interactive systems.}
  \label{fig:garden_path_cloud_erps}
\end{figure}

\section{Discussion and Outlook}

In this study we have developed a constructive, transparent, modular and parsimonious mapping from symbolic algorithms to neural networks. We first introduced a novel shift map, the versatile shift, that extends the generalized shift and allows for the real-time simulation of a range of symbolic models of computation. We then showed how VSs can be represented on a vectorial space through G\"odelization, obtaining piecewise affine-linear systems on the unit square known as nonlinear dynamical automata \citep{Tabor00a, TaborChoSzkudlarek13, graben2004language, graben2008towards}. Finally, we presented a modular R-ANN architecture that simulates the dynamics of NDA. The proposed architecture consists of three layers: a machine configuration layer representing the NDA state, and thus the symbolic data in the simulated automaton; a branch selection layer implementing the NDA switching rule, thus characterizing the automaton's decision space, or control; and the linear transformation layer implementing the set of piecewise affine-linear functions in the NDA, i.e. the vectorial representation of the symbolic operations defined in the transition table of the simulated automaton. Additionally, the linear transformation layer is itself modular, in that each operation specified by the $\delta$ transition function of the simulated automaton is applied by a  specific pair of units in the layer.

The mapping can be used to simulate any Turing machine through R-ANNs, thus making the architecture universal (an example of the mapping on Turing Machines is reported in the supplementary materials). In particular, it is possible to simulate the 7-states 4-symbols UTM by~\citet{minsky1962size} in real-time with a R-ANN consisting of 259 units \footnote{This implies a reduction factor of 1/3 when compared to the solution by \citet{siegelmann_math_1991, siegelmann_computational_1995}, which simulates Minsky's UTM with a network of 886 units.} (see \autoref{eq:n_units}), and the 6-states 4-symbols UTM by~\citet{neary_four_2009} with one consisting of 223 units. 

It is important to analyze some of the modeling choices that have been made in the R-ANN architecture we described. A choice worth discussing is that of implementing biases as synaptic projections from an always-active unit as opposed to implementing them as parameters intrinsic to the individual units. We decided for simplicity to add a bias unit. Nonetheless, a parameterized bias would have been equally reasonable. While it does not have strong bearings on the model here discussed, it is interesting to note that the specific choice of implementation does more or less put the emphasis on a predominantly synaptic computation versus a computation which is more distributed between the synaptic and the neuron level, reflecting similar issues to be considered in the biological domain.
A second consideration concerns the cell's boundaries in the NDA. In fact, the distance between the right bound of a cell and the left bound of the next one is zero. This poses some challenges, as even extremely small noise on the state vector at a boundary can lead to an erroneous application of the switching rule on the real state, and thus to a disruption of the computation. This is of course reflected in the dynamics of the associated R-ANN as well.
\citet*{siegelmann_computational_1995} solve this issue by using a Cantor encoding as opposed to a simple G\"{o}delization, ensuring a greater than zero distance between two encoded configurations with different leading symbols. The same methods can be applied here. Interestingly, by switching to a Cantor encoding, the Heaviside units in the BSL layer can be substituted with functionally equivalent Ramp units, so that the R-ANN would only make use of linear units \`a la Siegelmann and Sontag.

We will now first discuss the advantages of our approach over those based on eliminative connectionism, and then the advances that the present work brings to transparent connectionism.

Compared to eliminative approaches, our work allows the direct interpretation of the representations and the dynamics in the derived network in terms of symbolic computation. This has many important consequences. First, while conventional neural networks have to be trained on large data sets (usually using backpropagation or related algorithms, see \citealp{Werbos90}) our method does not require any training, as the synaptic weight matrix is explicitly designed from the machine table of the encoded automaton. Emergent representations and operations are not opaquely encoded in several hidden layers but transparently realized through G\"odelization of symbolic configurations.
Second, even when considering learning applications -- which we plan to explore in future developments -- the derived approach could bring about the exciting possibility of a symbolic read-out of a learned algorithm from the network weights; Note that in this architecture all weights are necessarily fixed, with the exception of the connections encoding the symbolic operations in the simulated automaton, i.e. those between the MCL and the LTL layer.
Third, anchoring the computation of the network to well-understood computation models is worthwhile when tackling problems that can benefit from the integration of the two perspectives. In the first example, we constructed a R-ANN ($24$ units) performing a FSM machine computation abstracting a CPG for animal locomotion. FSMs are widely used in locomotion controllers in robotics, because of their simplicity and strong theoretical grounding in relation to animal locomotion. On the other hand, neural implementations of CPG have many desirable characteristics (as discussed in \citealp{ijspeert2008central}) that are not present in FSM-based implementations, but they are difficult to engineer. We showed that by integrating the two approaches we can tackle the problem of pattern generation in robotic locomotion more effectively. Of course, a satisfactory solution would entail the use of continuous-time models in the mapping; nevertheless, our preliminary results already present distinct benefits in the integration of the two approaches as compared with their use in isolation.
Fourth, having a complete understanding of the network's inner workings allows for the intelligent manipulation of its parameters. In the discussed CPG example, understanding the computation carried out by the derived network allowed us to introduce a continuous control parameter eliciting a bifurcation in the dynamics of the network, as present in systems of coupled nonlinear oscillator models \citep{GolubitskyStewartEA99, GolubitskyStewartEA98, SchonerJiangKelso90, CollinsRichmond94}, widely studied in the CPG literature.

In regards to previous work on transparent connectionism, our work advances the field in several ways.
As a first advancement, by introducing VSs we are now able to use NDA to simulate a broad range of symbolic computation models in real-time, extending the original work by \citet{Moore90, Moore91a}. Interestingly, it would be straightforward to define $n$-sided infinite dotted sequences (where the dot splits a sequence in its $n$ one-sided infinite components), and extended VSs on these. By G\"odelization, we would obtain NDA on the $n$-dimensional hypercube, which could be simulated by R-ANNs through a straightforward extension of the architecture presented in this work. This would further extend the range of real-time simulable computational models to automata with multiple tapes or stacks \citep{Aho69, Weir94}.
Secondly, by basing our construction on NDA, we obtain an architecture characterized by a fully distributed representation coupled with a granular modularity, differentiating our approach from previous work and granting a series of advantages. The mapping is transparent not only with regards to the representations (the data), but also with regards to the symbolic operations defined in the simulated computational model and their control, all clearly localizable in the architecture. We regard this as an advance in itself (in line with the goals of transparent connectionism), but it also allows, for example, for the straightforward mapping of interactive automata networks to R-ANNs. This is of fundamental importance, as the framework of interactive computation provides a rich language for the description of many complex systems, for example in cognitive modeling. In the second example we constructed a network of interacting automata as a diagnosis and repair model \citep{lewis1998reanalysis, graben2004language, graben2008towards} for the reanalysis of linguistic garden path sentences. The network consisted of three PDA (two of them as TDRs), a VS, and one FSM as a master control program, with each component carrying out a specific and intelligible task in the overall computation. We then mapped this network to a R-ANN ($266$ units), thus obtaining a symbolic/connectionist implementation of a cognitive model. Interestingly, due to the multiple levels of hierarchical organization that can be present in the automata network (which comprises nesting, as in the diagnosis and repair network) and, thus, in the derived R-ANN, one could even speculate about thermodynamic limit networks when the number of modules approaches infinity, presenting emergent scale-free or small world properties  \citep{AlbertBarabasi02}. 
The granular modularity of our approach is also a key advancement when considering the possibility of correlational studies with neurophysiological measurements.
In previous work we showed how to devise large-scale biophysical observation models in order to correlate top-down modeling approaches with neurophysiological data obtained from bottom-up measurements \citep{Amari74, GrabenRodrigues13}. The process involves associating neural units of our model with neuronal masses \citep{SilvaHoecksEA74, JansenRit95} or Hebbian cell assemblies \citep{Hebb49, WennekersPalm09, Huyck09} in large-scale brain models, as investigated, e.g., in neural field theory. 
With this setup we then show that our observational models lead to improved interpretation, e.g of ``synthetic event-related brain potentials'' (as discussed in \autoref{sec:neural_observation_models}, see \citealp{graben2008towards, barres2013synthetic}) as used in computational neurolinguistics studies \citep{Gigley85, GrabenDrenhaus12, barres2013synthetic}, where mental/cognitive states can be associated to metastable states of a dynamical system.
In the second example presented here, we computed Amari's mean activation \citep{Amari74} as an observation model for the diagnose and repair R-ANN, in order to obtain synthetic ERPs \citep{graben2008towards, barres2013synthetic}. Qualitatively, the computed signal exhibited a similar divergence between conditions as measured in the experiment presented in \citet{FrischGrabenSchlesewsky04}.
While preliminary, these are already encouraging results for the development of our approach in this direction. In future work, we envisage that it will be possible to selectively correlate electrophysiological measurements with specific components in a derived R-ANN, as informed by a suitable symbolic model for the computation underlying the measured quantities.
As a third point of interest, the architecture presents a clear 2D spatial organization in its layout, particularly at the level of LTL (as highlighted in \autoref{fig:full_BSL}). In a NDA, different transformations are applied based on the position of the G\"odelized automaton data on the unit square. In the R-ANN architecture, this is implemented through the BSL, which performs a form of spatial pattern matching, activating a specific pair of units in the LTL through a lateral inhibition mechanism. When considering extensions to models of higher complexity, the functionality of BSL and LTL could be implemented through the use of a grid of units with receptive fields, as defined for example in self-organizing maps (SOMs, see \citealp{kohonen1982self, kohonen1998self}).

In future work, we plan to overcome fundamental issues with the current model which have bearing both in relation to learning applications and to the extension of the model to continuous dynamics.
For what concerns the learning of algorithms from data, the current model suffers from a missing end-to-end differentiability, due to the use of G\"odel encodings. This is a serious limitation, as it prevents the use of gradient descent methods for the training of the network's weights. Future work will have to address this limitation, possibly relying on methods of data access and manipulation akin to modern R-ANN approaches such as in \citet{weston2014memory, graves2014neural, grefenstette2015learning, joulin2015inferring, sukhbaatar2015end}.
Encouraging work on the learning of exponential state growth languages by Fractal Learning Neural Networks \citep{Tabor03, Tabor11} could also inform a revised trainable architecture.

With regards to the extension of the model to continuous dynamics, there are many ways in which this could be achieved in future work.
Importantly, we are mostly interested in extensions to continuous-time models that are excitable. In such systems, trajectories can be perturbed away from a stable equilibrium (or rest state) and come back to it only after a large excursion (or spike) in the phase space, upon sufficiently strong input; biophysical examples of excitable models were initiated in Hodgkin and Huxley, 1952.
One possibility would be to first extend the mapping to discrete-time excitable models (as in map-based neuronal models, see~\citealp{IbarzEA11,GirardiSchappoEA13}), and then move to continuous time via so-called \emph{suspension} procedures. There are some potential issues in this endeavor. First of all it would be crucial to first explore and understand the possible relationships between excitable regimes in neural models and symbolic dynamics in a computation. That is, to answer the question: how does the excitability property translate in the realm of symbolic computation? We think there could be meaningful answers to this question when tackled through the framework of interactive computation. Another potential issue is that the suspension process is non-unique and non-trivial in the general case; moreover, it does not guarantee that the excitability property will be preserved.

Excitability is a crucial matter when dealing with neural tissue of lower brain structures, such as the Brain stem, where it is possible to neurophysiologically identify clear and small neuronal networks. However, neural networks models are not the most appropriate level of description for higher cortical structures, due to the presence of large and highly interconnected neuronal masses. Models of these structures express slow but large scale processes as measured by LFP/EEG. In this context, an alternative approach to achieve continuous-time dynamics, which we have already explored to some extent in previous work, is by the framework of {\it heteroclinic dynamics}, where Turing machine configurations can be interpreted as metastable states with attracting and repelling directions \citep{graben2009inverse, Tsuda01, RabinovichEA08, krupa1997robust}, or by the framework of multiple-time scale dynamical systems \citep{desroches2013inflection, fernandez2015multiple}.



\section*{Acknowledgements}
This research has been supported by a Heisenberg fellowship (GR 3711/1-2) of the German Research Foundation (DFG) awarded to  PbG.





\end{document}